%% file: main.tex
\newif\ifextended

\extendedtrue

\ifextended
\documentclass[conference]{IEEEtran}
\IEEEoverridecommandlockouts
\else

\documentclass[year=23]{fmcad}
\IEEEoverridecommandlockouts
\fi

\def\BibTeX{{\rm B\kern-.05em{\sc i\kern-.025em b}\kern-.08em
    T\kern-.1667em\lower.7ex\hbox{E}\kern-.125emX}}

\usepackage{macros}
\usepackage{outlines}

\ifextended
\newcommand{\extref}[1]{#1}
\else
\newcommand{\extref}[1]{the extended version~\cite{wu2023lightweightext} of the paper}
\fi

\begin{document}

\title{Lightweight Online Learning for Sets of Related Problems in Automated Reasoning}

\makeatletter
\newcommand{\linebreakand}{%
  \end{@IEEEauthorhalign}
  \hfill\mbox{}\par
  \mbox{}\hfill\begin{@IEEEauthorhalign}
}
\makeatother

\author{\IEEEauthorblockN{Haoze Wu}
\IEEEauthorblockA{
\textit{Stanford University}\\
Stanford, CA, USA\\
haozewu@stanford.edu}
\and
\IEEEauthorblockN{Christopher Hahn}
\IEEEauthorblockA{
\textit{Stanford University}\\
Stanford, CA, USA\\
hahn@cs.stanford.edu}
\and
\IEEEauthorblockN{Florian Lonsing}
\IEEEauthorblockA{
\textit{Unaffiliated}\\
Linz, Austria \\
fml@florianlonsing.com}
\and
\IEEEauthorblockN{Makai Mann
%\thanks{The NASA University Leadership initiative (grant \#80NSSC20M0163) provided funds to assist the authors with their research, but this article solely reflects the opinions and conclusions of its authors and not any NASA entity.}
}
\IEEEauthorblockA{
\textit{MIT Lincoln Laboratory}\\
Lexington, MA, USA\\
makai.mann@ll.mit.edu}
\linebreakand
\IEEEauthorblockN{Raghuram Ramanujan}
\IEEEauthorblockA{
\textit{Davidson College}\\
Davidson, NC, USA \\
raramanujan@davidson.edu}
\and
\IEEEauthorblockN{Clark Barrett}
\IEEEauthorblockA{
\textit{Stanford University}\\
Stanford, CA, USA\\
barrett@cs.stanford.edu}
}
%haozewu@stanford.edu
%hahn@cs.stanford.edu
%makai.mann@ll.mit.edu
%fml@florianlonsing.com
%raramanujan@davidson.edu
%barrett@cs.stanford.edu

%\affil[1]{\orgdiv{Department of Computer Science}, \orgname{Stanford University}, \orgaddress{\state{CA}, \country{USA}}}

\maketitle

\input{00abstract}

\input{10intro}

\input{11related}
\input{30background}
\input{31sdcl}

\input{32impl}
\input{33instantiation}

\input{40experiments}
\input{50conclusion}

\bibliographystyle{IEEEtran}
\bibliography{bibli}

\ifextended
\input{90appendix}
\fi

\end{document}

%% file: 00abstract.tex
\begin{abstract}
We present Self-Driven Strategy Learning (\sys), a lightweight online learning methodology for automated reasoning tasks that involve solving a set of related problems.
\sys does not require offline training, but instead automatically constructs a dataset while solving earlier problems. It fits a machine learning model to this data which is then used to adjust the solving strategy for later problems.
We formally define the approach as a set of abstract transition rules.
We describe a concrete instance of the \sys calculus which uses conditional sampling for generating data and random forests as the underlying machine learning model.
We implement the approach on top of the \kissat{} solver and show that the combination of \kissat{}+\sys certifies larger bounds and finds more counter-examples than other state-of-the-art bounded model checking approaches on benchmarks obtained from the latest Hardware Model Checking Competition.
\end{abstract}

%% file: 10intro.tex
\section{Introduction}
\label{sec:intro}

Many automated reasoning tasks involve solving a set of related problems that share common structure. For example, in Bounded Model Checking~\cite{clarke2001bounded,biere2009bounded}, one repeatedly checks deeper and deeper unrolls of a transition system for a property violation. In iterative (e.g., counter-example-guided) abstraction refinement~\cite{clarke2000counterexample}, one verifies a condition on an increasingly precise model of a system. And in symbolic execution~\cite{king1976symbolic}, one analyzes the possible outcomes of a program on symbolic inputs by incrementally adding path conditions.  
Often, a fixed, predetermined high-level solving strategy (e.g., the choice of a solver and its parameter settings) is used in this iterative solving process. However, given the structural similarity within the set of problems, a natural question is: \emph{can we leverage information gathered while solving earlier problems to adjust the solving strategy for later problems on the fly?}

Adapting high-level solving strategies for particular problem distributions, a practice often termed \emph{meta-algorithmic design}~\cite{hoos2021automated}, is already a well-established technique. Automated configuration techniques~\cite{hutter2007boosting}, which optimize an algorithm's performance on a given set of problems, are widely used among practitioners. Per-instance algorithm selection techniques (e.g., SATzilla~\cite{xu2008satzilla}) train machine learning models to predict a suitable strategy for a given problem based on its structural characteristics. More recently, attempts to improve constraint solving using deep learning also generally follow the paradigm of choosing a particular problem distribution (which can be either broad, such as main-track benchmarks from SAT competitions~\cite{selsam2019guiding}, or narrow, such as graph coloring problems~\cite{yolcu2019learning}), gathering training data using instances in that distribution, and learning a strategy over the data. 

A shared, and arguably undesirable,trait of the aforementioned approaches is that they all involve an \emph{offline phase}, in which significant time and (often manual) effort are required to obtain an optimized solving strategy that can be used on new unseen problems. While the cost of the offline phase might be justified by the potential performance gain in the long run, the very distinction between an offline phase and an online phase already makes the reasoning less \emph{automated}.

Our first observation is that for an automated reasoning procedure whose execution involves solving a set $S$ of related problems, it is possible to move the meta-algorithmic design online, as part of the solving, by narrowing the scope of \emph{problem distribution} all the way down to $S$ itself. More concretely, we propose to solve some of the problems in $S$ not just once, but multiple times, each time using a different solving strategy from a space of candidates. The strategies used for solving later problems are selected based on information recorded during the multiple runs  (e.g., using lightweight machine learning techniques). We present this general method, which we term \textbf{S}elf-\textbf{D}riven \textbf{S}trategy \textbf{L}earning (\sys), as a set of transition rules, which can be used to model different ways of carrying out on-the-fly meta-algorithmic design.

Though there are many possible ways to instantiate \sys, we focus on a strategy space consisting of one fixed solver whose parameters are allowed to vary. One obvious method for exploring this strategy space involves choosing the first few problems, optimizing the parameter settings for them with a standard tuning procedure, and then using the optimized strategy for future problems. However, a drawback of this approach is that it only operates on a fixed set of problems and cannot explicitly take into account possible relationships between this set of problems and later problem instances.

To allow for such flexibility, our second observation is that a tuning procedure can be viewed, not as a way to select a specific solving strategy, but instead as a means of creating a dataset, where each data point is a pair consisting of a particular solving strategy and a particular problem in the problem set. 
A machine learning model is trained on this dataset to predict the effect of a given solving strategy on a given problem. The model can then be used as an oracle to select the solving strategy for future problems.

\begin{figure}[t]
\centering
       \includegraphics[width=0.95\linewidth]{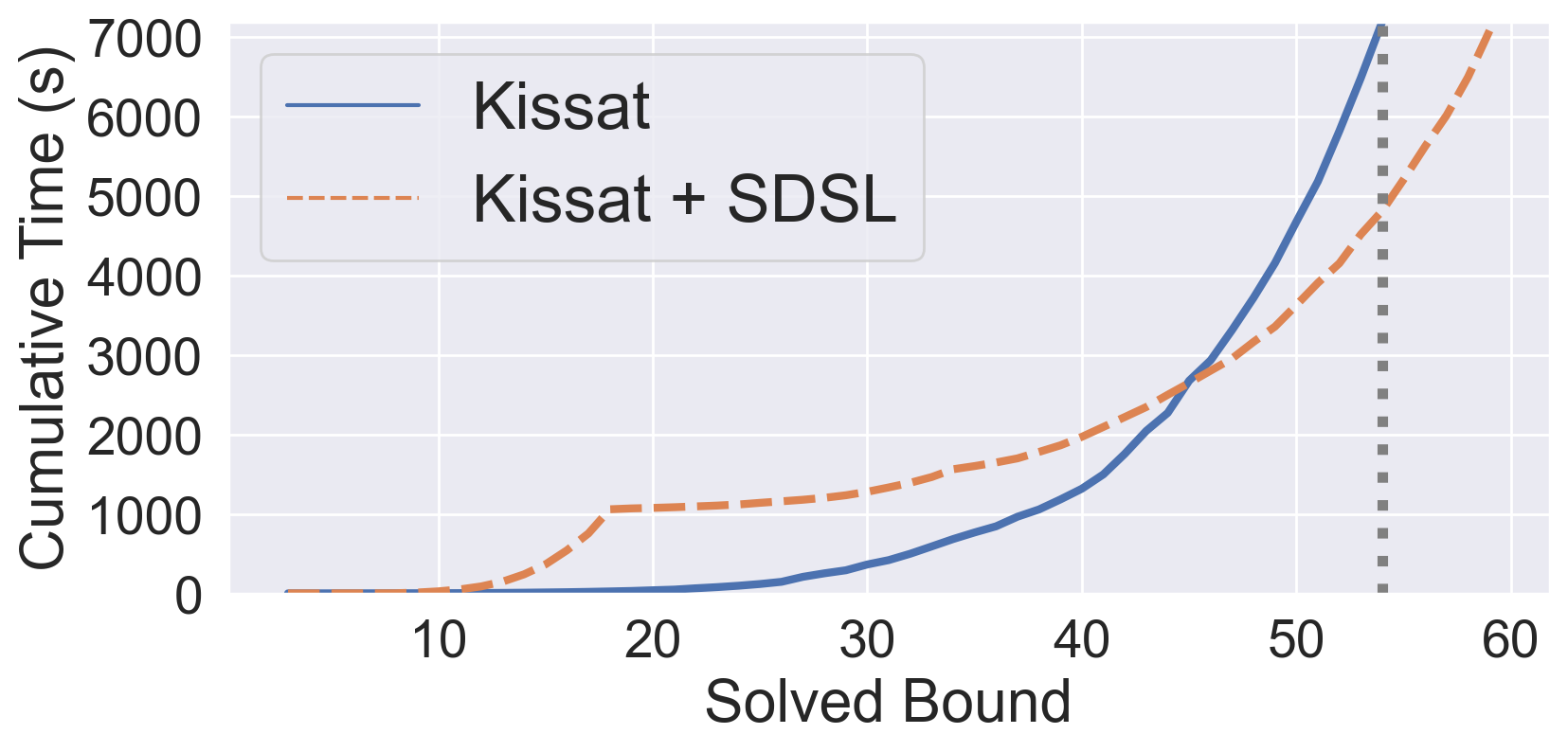}
       \caption{Executions of Bounded Model Checking with and without \sys on a hardware model checking benchmark (\benchmark{arb.n2.w128.d64}). The trajectories show the cumulative wall-clock time required to certify a bound.}
       \label{fig:sdsl-example}
\end{figure}

We apply our methodology to a case study of Bounded Model Checking (BMC) problems. We study different \sys instantiations and compare against existing model checkers. On satisfiable and unsolved bitvector benchmarks from the latest Hardware Model Checking Competition~\cite{preiner2020hardware}, our approach consistently boosts the performance of a BMC-procedure built on top of the \kissat~SAT solver~\cite{BiereFazekasFleuryHeisinger-SAT-Competition-2020-solvers}. Additionally, it compares favorably against state-of-the-art open-source model checkers \avr~\cite{goel2020avr} and \pono~\cite{mann2021pono}, contributing several unique solutions and speeding up many more. A preview on a single benchmark is shown in Fig.~\ref{fig:sdsl-example}. We see that \sys invests time learning a good solving strategy in the beginning, which results in better performance when solving later problems.

We summarize our main contributions as follows:
\begin{enumerate}
    \item we propose to move meta-algorithmic design online as part of solving a set of related problems;
    \item we propose a general methodology called Self-Driven Strategy Learning and present it formally as a set of transition rules; and
    \item we implement our approach and apply it to Bounded Model Checking problems, where it shows significant improvement over other state-of-the-art approaches.
\end{enumerate}

The rest of the paper is organized as follows. After a discussion of related work in Sec.~\ref{sec:rel}, we first define a basic calculus for iteratively solving a set of related problems in Sec.~\ref{sec:prelim}. Next, \sys is presented as an extension of this calculus with additional rules for data collection, learning, and strategy updates in Sec.~\ref{sec:method1}. We explore the design space of \sys in Sec.~\ref{sec:method2}, discussing how to sample training data and which machine learning models to use. In Sec.~\ref{sec:method3}, we describe in detail the instantiation of \sys for Bounded Model Checking. In Sec.~\ref{sec:exp} we present experimental results on Bounded Model Checking problems, and finally, we conclude with an account of current limitations and future directions in Sec.~\ref{sec:concl}.

%% file: 11related.tex
\section{Related Work}
\label{sec:rel}

Our approach is inspired and informed by several existing lines of work.

\paragraph{Incremental solving} A well-established paradigm for exploiting structural similarity is \emph{incremental solving}~\cite{hooker1993solving} in which each new query to a solver can be made by modifying the most recent formulas asserted in the previous query without resetting the solver. \sys is an orthogonal approach for leveraging structural similarity and may be preferable in cases where incremental solving is not beneficial or not supported.

In principle, the two can be combined. A straightforward way would be to switch to incremental solving mode after fixing a solving strategy. A tighter combination would require updating strategies \emph{between} incremental invocations, something that current solvers typically do not allow. In case one wants to both switch solvers on the fly and leverage incrementality, proof-transfer techniques such as solver state migration~\cite{biere2022migrating} are likely needed. For our particular BMC case study, we found that the direct use of an incremental SAT/SMT solver has mixed effects on performance (see \extref{App.~\ref{sec:incremental}}). This suggests that it might be worth revisiting BMC-specific incremental solving techniques such as conflict clause shifting~\cite{strichman2004accelerating} before investigating the interplay between \sys and incremental solving in BMC, which we leave for future work.

\paragraph{Automated Configuration} Our work is motivated by the success of offline meta-algorithmic design approaches such as automated configuration~\cite{hutter2007automatic,hutter2007boosting,khudabukhsh2016satenstein} and per-instance algorithm selection~\cite{xu2008satzilla,DBLP:journals/sttt/ScottNPNG23,xu2010hydra}. Automated configuration focuses on finding (near) optimal parameter settings of an algorithm for a fixed set of problems, using either local search or performance prediction techniques~\cite{hutter2014algorithm,ansotegui2015model,leyton2009empirical}. 
Per-instance algorithm selection techniques were among the first to utilize machine learning to improve constraint solving. The idea is to train an oracle to predict the performance (e.g., runtime) of a set of candidate algorithms on a formula based on its structural characteristics. \sys differs from both approaches primarily in that it moves meta-algorithmic design \emph{online}. 
%However, our instantiation of \sys draws inspiration from both automated configuration and per-instance algorithm selection: like the former, we probe the strategy space with stochastic search, and like the latter, we train a machine learning model to predict the performance of each solving strategy on each problem.

\paragraph{ML for AR} Machine Learning has been applied in multiple ways to expedite a variety of automated reasoning tasks, including satisfiability checking~\cite{selsam2019guiding,yolcu2019learning,balunovic2018learning,DBLP:conf/iclr/HahnSKRF21,DBLP:conf/sat/LiangGPC16,liang2018machine,wu2017improving,wu2019learning,you2019g2sat
}, Mixed-Integer Convex Programming~\cite{DBLP:journals/corr/abs-2012-13349, bertsimas-voice-of-opt}, program/function synthesis~\cite{golia2020manthan,golia2021engineering,parisotto2016neuro}, and symbolic execution~\cite{chen2018learning,10.1145/3460120.3484813}. 

While most existing techniques require an offline training phase, the general idea of using online learning also appears in previous work. The Conflict-Driven Clause Learning paradigm itself can be viewed as online learning. In the MapleSAT solver~\cite{DBLP:conf/sat/LiangGPC16,liang2018machine}, branching is formulated as a multi-armed bandit problem where the estimated reward of each arm (i.e., variable) is maintained and updated throughout the solving. This reinforcement learning interpretation of a dynamic branching heuristic is perhaps inspired by the study~\cite{liang2015understanding} of the popular VSIDS~\cite{moskewicz2001chaff} branching heuristic and its later variants~\cite{biere2015evaluating}, which also track a score for each variable during the solving. 
In contrast to this direction of online learning, \sys operates on a set of related problems rather than on a single instance. Moreover, \sys focuses on selecting from a set of existing strategies rather than inventing new ones.

To conclude this section, we remark that in practice, offline learning, ``in-solver'' online learning, and \sys could be combined to solve a set of related problems. For example, one could choose to use SAT/SMT solvers with built-in learning components, set the initial solving strategy using offline learning, and then use \sys to further customize the strategy online. The exploration of such combinations is beyond the scope of this paper, but is a promising future direction.
%CH: potential "dangerous" paragraph: not clear if sdsl really orthogonally improves when applying an offline learning technique a-priori (with the danger of sdsl being too expensive for too little gain).

%% file: 30background.tex
\section{Solving Sets of Related Problems}\label{sec:prelim}

%Many automated reasoning procedures involve iteratively solvinf a sets of related problems. An example is bounded model checking, where one checks whether a goal can be achieved by an execution over a transitional system that is within a fixed length \step. If no such execution exists, one checks again with a larger value $\step'$. This process is repeated until an execution that achieves the goal is found, or some pre-defined bound on execution length is reached. 
In this section, we present a simple calculus, \ua, for iteratively solving a set of related problems. Let $\formulas=\{\formula_1, \dots, \formula_\upperbound\}$ be a set of $\upperbound$ related formulas.
Assume we have a function $\funcEval: \formulas \times \strategyspace \rightarrow \{\sat,\unsat\}$, which takes as input a formula $\formula\in\formulas$ and a solving strategy \strategy (from a set \strategyspace called the \emph{strategy space}) and returns either \sat (satisfiable) or \unsat (unsatisfiable).  Additionally, assume that we can stop once any formula is \sat.

The rules of the basic \ua calculus are shown in Fig.~\ref{fig:rules-ua}. The rules operate over a \emph{configuration}, which is either one of the distinguished symbols \set{\success, \fail} or a tuple \tup{\step, \strategy}, where $\step\in[1,\upperbound]$ is the current formula index and $\strategy\in\strategyspace$ is the current solving strategy. The rules describe the conditions under which a certain configuration can transform into another configuration.
The \infrule{Next} rule says that if the current formula is unsatisfiable and the maximal index \upperbound has not been reached yet, then the current index will be increased. On the other hand, if the current formula is unsatisfiable and it is the last formula, the \infrule{Failure} rule transitions the system to the \fail configuration. The \infrule{Success} rule states that \success can be reached when the current formula is satisfiable.\unskip
\footnote{The conditions for progress and termination are inspired by BMC but are applicable in other settings when solving related problems.}

\begin{figure}[t]
    \centering
{
\small 
\begin{prooftree}
\AxiomC{$\step < \upperbound \quad \funcEval(\formula_\step, \strategy) = \unsat$}
\RightLabel{(\infrule{Next})}
\UnaryInfC{$\step, \strategy \Longrightarrow \step + 1, \strategy$}
\end{prooftree}
\begin{prooftree}
\AxiomC{$\step = \upperbound \quad 
\funcEval(\formula_\step, \strategy) = \unsat$}
\RightLabel{(\infrule{Failure})}
\UnaryInfC{$\step, \strategy \Longrightarrow \fail$}
\end{prooftree}
\begin{prooftree}
\AxiomC{$\funcEval(\formula_\step, \strategy) = \sat$}
\RightLabel{(\infrule{Success})}
\UnaryInfC{$\step, \strategy \Longrightarrow \success$}
\end{prooftree}
}
\caption{Transition rules for solving a set of related problems. The starting configuration is \tup{1, \strategy}.} 
\label{fig:rules-ua}
\end{figure}

An \ua-\emph{execution} is a sequence of configurations that respect the rules in \ua.
Note that no rule updates the solving strategy \strategy.  We augment the calculus with strategy updates next.

Given two configurations $C, C'$, we use $C \provable C'$ to denote $C$ can transition (in one or more steps) to  $C'$. We state the following two propositions which are straightforward to verify.
\begin{proposition}[Soundness and Completeness]
\formulas contains a satisfiable formula if and only if $\tup{1, \strategy} \provable \success$. 
\label{prop:sound-complete}
\end{proposition}
\begin{proposition}[Termination]
There exist no infinite \ua-executions.
\label{prop:terminate}
\end{proposition}

%% file: 31sdcl.tex
\section{Self-Driven Strategy Learning}
\label{sec:method1}

\begin{figure}[t]
    \centering
{
\small
\begin{prooftree}
\AxiomC{
$\strategy_\samp\in\strategyspace \quad 
\stepnew \leq \step  \quad
\cost = \funcCost(\formula_\stepnew, \strategy_\samp)
$
}
\RightLabel{(\infrule{Collect})}
\UnaryInfC{
$\step, \strategy, \dataset, \oracle 
\Longrightarrow 
\step, \strategy, \dataset \cup \set{\tup{\strategy_\samp, \stepnew,\cost}}, \oracle$
}
\end{prooftree}
\begin{prooftree}
\AxiomC{$\oracle' = \funcFit(\dataset)$}
\RightLabel{(\infrule{Train})}
\UnaryInfC{
$\step, \strategy, \dataset, \oracle
\Longrightarrow 
\step, \strategy, \dataset, \oracle'
$
}
\end{prooftree}
\begin{prooftree}
\AxiomC{
$\strategyspace_\samp\subseteq \strategyspace \quad
\strategy' = \argmin_{v_\samp\in \strategyspace_\samp}\oracle(v_\samp, \step) 
$
}
\RightLabel{(\infrule{Strategize})}
\UnaryInfC{
$\step, \strategy, \dataset, \oracle
\Longrightarrow 
\step, \strategy', \dataset, \oracle
$
}
\end{prooftree}
}
\caption{Additional transition rules for Self-Driven Strategy Learning.} 
\label{fig:rules-sdsl}
\end{figure} 

\subsection{Informal Presentation} 
\label{subsec:informal}

Self-Driven Strategy Learning (\sys) attempts to learn, on the fly, which solving strategy among a set of candidates $\strategyspace$ to use for each formula in $\formulas$. Learning is based on data gathered during solving. To obtain the data, we occasionally solve a formula multiple times, each time with a different strategy in \strategyspace. For a strategy $\strategy_\samp$, we record its effect, $\cost\in\R$, when solving $\formula_\step$ by creating a data point \tup{\strategy_\samp, \step, \cost}, where \cost is a measure of the cost of the strategy. For example, \cost could be the total run time required to solve $\formula_i$.

Given such a dataset, an oracle $\oracle : \strategyspace\times [1,\upperbound]\mapsto \R$ is trained to predict the cost of a given strategy when run on a given formula. When solving a new formula, we select the one that \oracle predicts will be most effective, and as more data is collected with each call to $\funcEval$, \oracle is updated.

An essential characteristic of \sys is that the training data is gathered for a specific, and \emph{a priori} unknown, set of formulas in an \emph{online} and \emph{automatic} manner, as part of the solving process. This approach has two challenging implications. The learning process must not incur a large overhead; otherwise, insufficient time is left for actual solving.
Additionally, the choice of \strategyspace is crucial as it must be large enough to contain good candidate strategies but also not too large to explore. We address these challenges in Sections~\ref{sec:method2} and \ref{sec:method3}, respectively.

\subsection{Formal presentation}
\label{subsec:formal}

Formally, we present \sys as an extension of \ua.  Configurations are as in \ua except that tuple configurations \tup{\step, \strategy, \dataset, \oracle} have two additional components (assumed to be left unchanged by rules in \ua):
$\dataset\in \powerset{\strategyspace\times[1,\upperbound]\times\R}$ is a dataset, each of whose members records the result of running a single strategy on a single formula; and $\oracle : \strategyspace\times [1,\upperbound]\mapsto \R$ is an oracle (e.g., a machine learning model) that predicts the cost of a strategy on a formula. Initially, \dataset is empty, and \oracle is arbitrary (e.g., always return 0). The additional transition rules of \sys are described in Fig.~\ref{fig:rules-sdsl}.

The \infrule{Collect} rule samples a strategy $\strategy_\samp\in\strategyspace$, evaluates its cost when solving $\formula_\stepnew$, and augments \dataset with this new data. The rule is parameterized by a function $\funcCost : \formulas\times\strategyspace\mapsto \R$. The \infrule{Train} rule updates the oracle \oracle with a new one trained on the current dataset \dataset. It is parameterized by a machine learning algorithm \funcFit (e.g., k-NN, tree ensemble, deep learning, etc.). Finally, the \infrule{Strategize} rule updates the current strategy by sampling a set of strategies $\strategyspace_\samp$ from \strategyspace and choosing the one with the best predicted cost for the current index \step. The extended calculus is still sound and complete (i.e., Proposition~\ref{prop:sound-complete} still holds). Since the added rules can effectively be applied at any time, Proposition~\ref{prop:terminate} only holds if we allow only a finite number of applications of the new rules.

Note that in the \infrule{Collect} rule, the results of solving the formula $\formula_\stepnew$ are discarded, as $\formula_\stepnew$ must have been solved already in some previous application of the \infrule{Next} rule. It is possible to extend the SR calculus to allow \unknown results from $\funcEval$, but the completeness property would be lost.

A reasonable strategy for applying \sys rules is as follows: 
\begin{enumerate}
    \item After every application of \infrule{Next}, issue one \emph{learning epoch}; that is, apply \infrule{Collect} $\paramnumsamples$ times on the current problem, then apply \infrule{Train}; 
    \item If \infrule{Train} has been applied at least once, apply \infrule{Strategize} whenever $\step$ is updated;
    \item If the \emph{estimated learning time} exceeds some threshold $\paramstopthreshold$, override the first policy and do not issue any more learning epochs; 
    \item Terminate whenever \infrule{Success} or \infrule{Failure} applies. 
\end{enumerate} 

The estimated learning time is calculated as the time spent on learning so far plus $\paramnumsamples \cdot t$, where $t$ is the runtime of solving the current problem using the current strategy. If $|\strategyspace|$ is small, it may be reasonable to use $\paramnumsamples=|\strategyspace|$ and try each strategy from \strategyspace. In the more general scenario where $\paramnumsamples \ll |\strategyspace|$, the choice of which samples to use impacts the quality of the dataset. We discuss this choice and present a conditional sampling procedure in Sec.~\ref{subsec:cond-sampling}.
The purpose of restricting the training time in step three is to ensure that training does not dominate the total time taken.
This simple criterion for when to stop learning already works reasonably well in practice. We leave the exploration of more sophisticated heuristics to future work.

%% file: 32impl.tex
\section{Design Space in \sys}
\label{sec:method2}

This section discusses the design space in the implementation of \sys and proposes solutions to the following questions: 
\begin{enumerate*} [1) ]
    \item How should training data be sampled?
    \item Which machine learning model and training algorithm should be used?
\end{enumerate*}
The solutions we propose focus on the case where a strategy is simply a set of values for a specific set of solver parameters. In this case, the strategy space is the cartesian product of $\numfeatures$ sub-strategy spaces, each representing a single parameter: $\strategyspace = \strategyspace_1 \times \ldots \times \strategyspace_\numfeatures$. The set of possible values for each parameter can vary (e.g., parameter values could be Booleans, strings, or numbers), but for now we assume each $\strategyspace_i$ is finite.

\subsection{Gathering Informative Training Data}
\label{subsec:cond-sampling}

To make the most informed decision, we could try all candidate strategies on all previously considered problems, but this is infeasible when $|\strategyspace|$ is large.
In the following, we consider the scenario where \paramnumsamples samples are drawn from a strategy space \strategyspace, where $\paramnumsamples \ll |\strategyspace|$. 

In this restrictive setting, we need to ensure our dataset contains a sufficient number of low-cost strategies (if there are any). Sampling uniformly is unlikely to achieve this goal because in practice, many or most candidate strategies could have high cost. For this reason, we propose to \emph{explicitly} favor low-cost strategies in the sampling process. One way to do this is by using Markov-Chain Monte-Carlo (MCMC) sampling, which in our setting can be used to generate a sequence of solving strategies with the desirable property that in the limit, strategies with the lowest cost are most frequently drawn. A popular MCMC method is the Metropolis-Hastings (M-H) Algorithm~\cite{mh}, instantiated in the context of \sys as follows: 
\begin{enumerate}
    \item Choose a current strategy \strategy;
    \item Propose to replace the current strategy with a new one $\strategy'$, which comes from a \emph{proposal distribution} $q(\strategy' | \strategy)$;
    \item If $\funcCost(\formula, \strategy') \leq \funcCost(\formula, \strategy)$, accept $\strategy'$ as the current strategy;
    \item Otherwise, accept $\strategy'$ as the current strategy with some probability $a(\strategy{\rightarrow}\strategy')$ (e.g., a probability inversely proportional to the increase in cost);
    \item Go to step 2.
\end{enumerate}
This process is repeated until \paramnumsamples samples are drawn. Importantly, under this scheme, a proposal that results in lower cost is always accepted, while a proposal that does not may still be accepted. This means that the algorithm greedily moves to a better strategy whenever possible, but also has a means for escaping local minima. 
In our implementation, the acceptance probability is computed using a common method~\cite{mcmc} described as follows. We first transform $\funcCost(\formula,\strategy)$ into a probability distribution $p(\strategy)$: 
\[
p(\strategy) \propto \exp(-\beta \cdot \funcCost(\formula,\strategy)) \enspace,
\]
where $\beta > 0$ is a configurable parameter. The acceptance probability is then computed as:
\begin{align*}
a(\strategy{\rightarrow}\strategy') &= \min\left(1, \frac{p\left(\strategy'\right)}{p\left(\strategy\right)}\right)  \\
&= \min\left(1, \exp\left(\beta \cdot \left(\cost - \cost'\right)\right)\right) \enspace,
\end{align*}
where $\cost = \funcCost(\formula,\strategy)$ and $\cost' = \funcCost(\formula,\strategy')$. Under this acceptance probability, the larger that $\cost'$ is compared to \cost, the lower the probability to accept. On the other hand, the larger $\beta$ is, the more reluctant we are to move to a worse proposal.

To ensure the aforementioned convergence property of MCMC in the limit, the proposal distribution must be both \emph{symmetric} and \emph{ergodic}.\footnote{
A proposal distribution $q$ is symmetric if $q(\strategy' | \strategy) = q(\strategy | \strategy')$ for any $\strategy, \strategy'\in\strategyspace$
and is ergodic if there is a non-zero probability of reaching a strategy $\strategy\in\strategyspace$ from any other strategy $\strategy'\in\strategyspace$ in a finite number of steps.
}
For discrete search spaces, a common proposal distribution is the symmetric random walk, which moves to one of the \emph{neighbors} of the current sample with equal probability. For our strategy space, we define the neighbors of a strategy as all strategies for which exactly $k$ parameter values are different. We use $k=1$ in our implementation.

Note that MCMC sampling can be used not only in the data collection process, but also in the \infrule{Strategize} rule (i.e., to choose $\strategyspace_\samp\subseteq \strategyspace$). Since in this case we use the machine learning model as an oracle of \funcCost (which is much cheaper than calling a solver), a larger sample size is affordable. 

The sampling scheme presented above largely coincides with many local search approaches used in the automated configuration literature~\cite{hoos2021automated}. Borrowing more insights from that literature and devising more sophisticated sampling schemes are interesting directions for future work.

\subsection{Lightweight Online Learning}
\label{subsec:learn}

%n = number of training examples, m = number of features, n' = number of support vectors,
%k' = number of trees

%TODO: do complexity analysis for sdsl BMC? in Section VI?

%Decision Tree

%Train Time Complexity=O(n*log(n)*m)
%Test Time Complexity=O(m)
%Space Complexity = O(depth of tree)

%Random Forest

%Train Time Complexity=O(k'*n*log(n)*m)
%Test Time Complexity=O(m*k')
%Space Complexity = O(k'*depth of tree)

In the online setting, the machine learning model must generalize from sparse data in limited time.
This means the model needs to be both robust against outliers and efficient to train. Training a neural network from scratch, for example, is likely unsuitable, because it requires large amounts of data and, depending on the architecture, could be costly to train. On the other hand, lightweight ensemble models, which consist of a set of sub-models with different strengths and weaknesses, are well-suited for \sys.

Our data is what is often called \emph{tabular data}, that is, it can be represented as a table with rows and columns, where each row corresponds to a sample, and each column corresponds to a feature. When the strategy space consists of parameter settings, each sample has $\numfeatures + 2$ features: the \numfeatures parameters, the problem index, and the cost. Tree-based ensemble methods such as \emph{random forests} are generally considered to be a good match for such data~\cite{breiman2001random}.

A random forest consists of a set of $B$ \emph{regression trees} $\{f_1,\ldots,f_B\}$. Each tree is trained independently by sampling data from the data set \dataset. A regression tree makes a prediction by following a path from the root node to a leaf node, based on the values of the input features, and returning the cost associated with the leaf node (which generally is the average of the costs of the training points that map to that leaf node). The predictions of a random forest $f$ are computed by averaging the predictions of the individual trees in $f$.

A random forest is both efficient to train and efficient for prediction~\cite{louppe2014understanding}. The time complexity of training a random forest with $B$ trees is $O(B\cdot m\cdot n \cdot \log m)$, where $m$ is the number of data points and $n$ is the number of input features. The inference time complexity for a random forest is $O(B\cdot n)$. 
 
Many machine learning algorithms are themselves parameterized and the performance of the model depends on a good choice of the hyperparameters. For a tree-based algorithm like random forest, an important hyperparameter is the maximal depth allowed for each individual regression tree: too shallow, and the model's prediction will be inaccurate; too deep, and the model might overfit to outliers.\unskip
\footnote{
Instead of tuning the tree depth while fixing the number of trees, one could alternatively grow deep trees and tune the number of trees (to be large enough). However, this makes training and prediction much more costly.
}
The standard way to find suitable values of the hyperparameters is via (cross\nobreakdash-)validation: split the data into a training set and a validation set, train models with different hyperparameters on the training set, evaluate them on the validation set, and pick the best one. However, in the \sys setting where data is already sparse, validation is less feasible because it is hard to make sure that both the training set and the validation set are representative of the input space. Instead, we propose the following pragmatic heuristic: start by training a random forest with shallow trees and then retrain with incrementally deeper trees as needed until the training score is high enough.

%The benefit of using a random forest in the \sys calculus is two-fold: 1) individual trees in the forest can be kept shallow, resulting in an efficient training; using a single deep decision tree on the other hand often results in overfitting, and 2) The size of the forest can be adjusted depending on the training budget, making it a flexible approach with a small number of hyperparameters.
%A downside of using a random forest is it's black-box nature. If one wishes to have a white-box approach, one can rely on a single decision tree, which also results in significant gains applied in the \sys setting (see Appendix for an ablation).
%

%% file: 33instantiation.tex
\section{Case study: Bounded Model Checking}
\label{sec:method3}

Bounded Model Checking (BMC)~\cite{clarke2001bounded,biere2009bounded,strichman2004accelerating} is a well-known technique for checking whether a property $P$ holds along bounded executions of a given system $M$.  The algorithm starts by checking all executions of length $k$; if no counter-example is found, $k$ is increased and the system checked until either a counter-example is found, the problem becomes intractable, or some upper bound on $k$ is exceeded.

BMC is useful in practice for at least two reasons. First, it is often the most efficient way to find counter-examples (if they exist) when trying to prove that a system has a particular property.
Second, when techniques capable of providing a full (i.e. unbounded) proof fail (which is often the case in practice), BMC still establishes a certain confidence in the system by providing formal guarantees for bounded executions. The larger the \emph{certified bound}, the stronger the guarantee.

%This problem can be efficiently reduced to a satisfiability problem (e.g. modulo propositional logic, theories of bit-vectors, and/or arrays) and can therefore be solved by SAT/SMT solvers. 
A basic BMC formula for checking whether a property $P$ holds for a system $M$ along executions of length $k$ is:
\begin{equation*}
I_0 \land \bigwedge^{\bmcbound-1}_{\bmccycle=0} \rho(\bmccycle, \bmccycle+1) \land (\bigvee^\bmcbound_{\bmccycle=0} \neg P_\bmccycle)    \enspace ,
\end{equation*}
where $I_0$ represents the initial state of $M$, $\rho(\bmccycle, \bmccycle + 1)$ represents how the system evolves in a single step, and $P_\bmccycle$ represents the property at step $\bmccycle$ in the execution. This formula is satisfiable iff there is an execution of length less than or equal to $\bmcbound$ such that the property $P$ does not hold at the end of the execution.

In practice, when the bound is increased, additional constraints are added stating that previously checked states are safe (in order to prune the search space). For example, suppose the check for bound $k'$ is unsatisfiable.  To check bound $\bmcbound > \bmcbound'$, we use the following formula:
\begin{equation*}
\bmcproblem(k',k) := I_0 \land \bigwedge^{k-1}_{i=0} \rho(i, i+1) \land {\color{red}\bigwedge^{k'}_{i=0} P_i} \land  (\bigvee^k_{{\color{red}i=k'+1}} \neg P_i) \enspace .
\label{eq:bmc}
\end{equation*}

We use BMC as a case study for our approach. For a given system and property, we solve the following set of problems:
\[
\formulas = \set{\bmcproblem(\bmcbound - \bmcstepsize,\bmcbound)\ | \ \bmcbound = \bmccycle \cdot \bmcstepsize, 1 \leq \bmccycle \leq \upperbound} \enspace ,
\]
where $\bmcstepsize$ is the \emph{step size}.
We focus on hardware model checking problems where the set of formulas to solve is in the theory of bitvectors~\cite{barrett2010satisfiability}. We use standard techniques to encode the bitvector problems as Boolean satisfiability (SAT) problems~\cite{kroening2016decision}. Thus, $\formulas$ is a set of Boolean formulas, and we can implement $\funcEval$ using an off-the-shelf SAT solver. We use the state-of-the-art \kissat SAT solver~\cite{BiereFazekasFleuryHeisinger-SAT-Competition-2020-solvers}.

In the following, we discuss the choice of the cost function $\funcCost$ and the strategy space \strategyspace for this case study.

\subsection{Choosing the Cost Function}

One plausible cost function for a strategy \strategy and a formula $\formula_i$ is the ratio of the runtime to that of some default strategy $\strategy_0$, i.e., if the runtime is $t$ with strategy $\strategy$ and $t_0$ with $\strategy_0$, then $\funcCost(\formula_i, \strategy)=\frac{t}{t_0}$. While this definition works in practice, the use of runtime makes \sys's behavior non-deterministic across different runs. This is undesirable for many reasons, including experimental reproducibility. Therefore, we instead use the \emph{number of conflicts} generated by the SAT solver, which is accepted as a good proxy for runtime~\cite{beskyd2023domain}. Given the same parameter settings, the number of conflicts generated by \kissat on the same problem is deterministic.

\subsection{Choosing the strategy space}
\label{subsec:strategyspace}

As discussed in Sec.~\ref{subsec:formal}, the choice of the strategy space is crucial to the effectiveness of \sys. \kissat has over 90 configurable parameters, so considering all of them is impractical. One plausible approach is to rely on expert/domain knowledge and empirical studies to identify a reasonable set of parameters to consider. We follow this approach to define two strategy spaces for \kissat.

The first one, \strategyspaceexp (Table~\ref{tab:strategyspace-exp}), is based on a study by Dutertre~\cite{dutertre2020empirical} on the effect of SAT solver parameters on bitvector problems.\unskip 
\footnote{
We consider all options considered in Table 2 of \cite{dutertre2020empirical}, except four: ``lucky'' and ``walk'' control procedures that find satisfying assignments independent of the main search; ``scan-index'' is not available in \kissat; and ``compacting'' is a data-structure optimization that we do not believe has strong correlations with the number of conflicts. Noting that \cite{dutertre2020empirical} does not consider any options related to branching, we additionally consider the \opt{bumpreasonsrate} parameter, which controls the eagerness of reason-side literal bumping~\cite{DBLP:conf/sat/LiangGPC16} and reportedly~\cite{BiereFazekasFleuryHeisinger-SAT-Competition-2020-solvers,biere2017cadical} has significant impact on SAT Competition benchmarks. 
}
We allow two possible values for each parameter, the default one and an alternative one. For options that were found to be beneficial in \cite{dutertre2020empirical}, we include the corresponding parameter in \kissat and for non-Boolean parameters, we set the alternative value to be more aggressive;\unskip
\footnote{The alternative values are selected as follows: \opt{*int} parameters are divided by 10; \opt{*lim} parameters are divided by 100; and \opt{*effort} parameters are doubled.  This works well in practice, and in further testing, setting the parameters to other reasonable values did not significantly alter the overall results.
In the future, it might be advisable to obtain expert knowledge also on the specific values of the parameters.
} 
for Boolean parameters, we simply set the alternative value to be the opposite of the default.
In total, \strategyspaceexp contains $8192$ $(2^{13})$ possible parameter settings.

\begin{table}[t]
\setlength\tabcolsep{8pt}
\centering
\caption{The strategy space \strategyspaceexp based on ~\cite{dutertre2020empirical}. } 
\begin{tabular}{ccc}
\toprule
\kissat option & default & alternative\\
\midrule
\opt{and} & 1 & 0\\
\opt{bumpreasonsrate} & 10 & 1 \\
\opt{chrono} & 1 & 0 \\
\opt{eliminateint}  & 500 & 50   \\ 
\opt{eliminateocclim}  & 2000 & 20   \\ 
\opt{forwardeffort}  & 100 & 200   \\ 
\opt{ifthenelse}  & 1 & 0   \\ 
\opt{probeint}  & 100 & 10   \\ 
\opt{rephaseint}  & 1000 & 100   \\
\opt{stable}  & 1 & 0   \\ 
\opt{substituteeffort}  & 10 & 20   \\ 
\opt{subsumeocclim}  & 1000 & 10   \\ 
\opt{vivifyeffort}  & 100 & 200   \\ 
\bottomrule
\end{tabular}
\label{tab:strategyspace-exp}
\end{table}

The second strategy space, \strategyspacedev (Tab.~\ref{tab:strategyspace-dev}), is based on suggestions made by the developer of \kissat.\unskip
\footnote{See \url{https://github.com/arminbiere/kissat/issues/25}} 
It contains significantly fewer possible parameter settings (216).

\begin{table}[ht]
\setlength\tabcolsep{8pt}
\centering
\caption{The strategy space \strategyspacedev.} 
\begin{tabular}{ccc}
\toprule
\kissat option & default & alternative(s)\\
\midrule
chrono & 1 & 0\\
phase & 1 & 0 \\
stable & 1 & 0, 2 \\
target  & 1 & 0, 2   \\ 
tier1  & 2 & 1  \\ 
tier2  & 6 & 3, 9  \\ 
\bottomrule
\end{tabular}
\label{tab:strategyspace-dev}
\end{table}

Designing principled ways to automatically construct the strategy space (e.g., using techniques for assessing parameter importance~\cite{hutter2014efficient}) is an important direction for future work.

\subsection{Implementation}
\label{subsec:bmcimplementation}

We implemented an \sys-based BMC procedure in \textsc{Python3}.\unskip
\footnote{Available at \url{https://github.com/anwu1219/sdsl/}}
Our prototype takes as input a model checking problem in the \solver{btor}/\solver{btor2} format~\cite{brummayer2008btor,niemetz2018btor2} and can run BMC on that input with or without \sys. We implemented \sys following the strategy described in Sec.~\ref{subsec:formal}.
The BMC step size and the maximal bound are also command-line arguments. Additional input arguments include: 
\begin{enumerate} [1) ]
    \item \strategyspace: path to a \solver{csv} file representation of the strategy space (e.g., Tabs.~\ref{tab:strategyspace-exp} and ~\ref{tab:strategyspace-dev});
    \item \paramstopthreshold: the time budget for the learning epochs (see Sec.~\ref{subsec:formal}), by default 15\% of the total time limit;
    \item \paramnumsamples: the number of samples to draw per learning epoch, by default 100;
    \item The number of samples to draw in the \infrule{Strategize} rule, by default 500;
    \item The number of trees in the random forest, by default 50;
    \item The initial tree depth, by default a third of the number of parameters in \strategyspace;
    \item The random seed, by default 0.
\end{enumerate}
The default values are used in all experiments unless otherwise specified.

The formula $\bmcproblem(k',k)$ is generated online by first creating a bitvector formula using the \pono Model Checker~\cite{mann2021pono}, then bit-blasting it into a SAT formula using the \boolector solver~\cite{brummayer2009boolector}. The versions of the solvers are reported in \extref{App.~\ref{sec:versions}}. Our prototype does not leverage incrementality for reasons discussed in Sec.~\ref{sec:rel}.
We use the Scikit-Learn machine learning library~\cite{pedregosa2011scikit} for training the Random Forest. Apart from the number of trees and the depth of the trees, we use the default hyperparameters of Scikit-Learn's Random Forest module. The prototype runs on one thread, though the sampling, training, and inference are in principle parallelizable.

%% file: 40experiments.tex
\section{Experimental Evaluation}
\label{sec:exp}

\input{results/unsolved_sdsl_vs_kissat}

We consider the bitvector track benchmarks from the latest Hardware Model Checking Competition (HWMCC)~\cite{preiner2020hardware}. We omit all unsatisfiable benchmarks, since these are not solvable using BMC. What remain are 65 benchmarks that were reported to be satisfiable during the competition and 24 benchmarks that were unsolved during the competition. All experiments are performed on a cluster equipped with Intel(R) Xeon(R) CPU E5-2637 v4 @ 3.50GHz running Ubuntu 20.04. Each job is given one physical core and 8 GB memory.

\subsection{Unrolling the unsolved benchmarks}

In the first experiment, we focus on the 24 unsolved benchmarks. For each benchmark, the goal is to either find a property violation or to prove that the property holds for as large a bound as possible. A CPU time limit of 2 hours is given for each benchmark. We consider two BMC step sizes: 1 and 10.\unskip
\footnote{The value of 10 is chosen based on a study by Lonsing~\cite{centaur-talk}.}
For each step size, we run as baselines our \kissat-based BMC implementation without \sys (denoted \kissat) and the BMC engine of \pono (denoted \pono), which makes incremental calls to \boolector to solve bitvector queries.

\subsubsection{Performance of \sys using the strategy space in Tab.~\ref{tab:strategyspace-exp}}
We first evaluate \bmcsdslexp, the \sys-extended BMC procedure using \strategyspaceexp (Tab.~\ref{tab:strategyspace-exp}) as the strategy space. The results are shown in Tab.~\ref{tab:eval-exp}. We report the largest \emph{solved} (i.e., certified or falsified) bound \colbound for each configuration. For \bmcsdslexp and \kissat, we also show the total time to solve all formulas up until the largest commonly solved bound (\coltimecommonstep). For \bmcsdslexp, this includes the time spent on learning. We further report the number of learning epochs (\coliters) and the time spent on learning (\coltimetrain) for \bmcsdslexp. Graphic illustrations in the style of Fig.~\ref{fig:sdsl-example} and the duration of each training epoch are presented in \extref{App.~\ref{sec:raw} and App.~\ref{sec:training-ep}, respectively}.

When the BMC step size is 1, \bmcsdslexp is able to certify larger bounds compared with the baseline configurations on 22 out of the 24 benchmarks, with an average bound increase of $3.9$ ($52.6-48.7$). This improvement is highly non-trivial, considering that to reach a larger bound, \bmcsdslexp needs to 1) certify all the formulas up to the baseline bound; 2) spend time (on average 975 seconds) learning a solving strategy; and 3) solve an additional set of harder formulas with the remaining time, one for each increase in the bound size.
Comparing \coltimecommonstep sheds further light on the performance gain enabled by \sys: on average, \bmcsdslexp is $1.3\times$ faster ($\frac{6354}{4811}$) on the set of commonly solved problems. The fraction of \coltimecommonstep that \bmcsdslexp spends on actual solving (not including learning) is $3836$ seconds ($4811 - 975$). Thus, on average a $1.7\times$ speedup ($\frac{6354}{3836}$) is achieved in the sheer performance of the SAT solver. Upon closer examination, the learning time is dominated by the data collection, with actual training and inference only taking 2.1\% of \coltimetrain on average.

When using BMC step size 10, both the \kissat-based baseline and \bmcsdslexp find counter-examples on 8 benchmarks (highlighted in red). In all but one of those benchmarks, \bmcsdslexp finds counter-examples faster. Additionally, \bmcsdslexp certifies a larger bound than \kissat on 4 benchmarks. For the remaining 12 benchmarks, the two configurations certify the same bounds, but \bmcsdslexp reduces the runtime on only 3 of them. One explanation for this is that the number of affordable learning epochs is significantly smaller when the step size is 10 due to the increased hardness of individual formulas. 
As a result, fewer strategies are considered. For example, on \benchmark{arb.n2.w128.d64}, a total of 871 unique solving strategies are evaluated when the step size is 1, whereas only 175 strategies are evaluated when the step size is 10.
Nonetheless, overall, \bmcsdslexp is still $1.3\times$ faster ($\frac{3712}{2927}$) at certifying the same bounds.

It is important to note that using step size 10 does not necessarily lead to a larger certified bound. Take \benchmark{circ.w128.d128} for example: \bmcsdslexp can unroll to an execution length of 46 with step size 1 while only unrolling to 30 with step size 10. This also applies to an incremental solver like \pono, which certifies an execution length of 39 with step size 1 versus 20 with step size 10.
This suggests that the optimal step size varies in practice.

\input{results/unsolved_sdsl_vs_kissat_biere}

\subsubsection{Performance of \sys using the strategy space in Tab.~\ref{tab:strategyspace-dev}}
We repeat the same experiment for the other \sys configuration \bmcsdsldev, which uses the smaller strategy space \strategyspacedev (Tab.~\ref{tab:strategyspace-dev}). The result is shown in Tab.~\ref{tab:eval-dev}. To summarize, \bmcsdsldev still boosts the performance of \kissat though the overall gain is less. For step size 1, the average solved bound by \bmcsdsldev is 49.4 compared to 48.7 by \kissat. The overall reduction in \coltimecommonstep is not significant (2.8\%) though the reduction in the pure solving time (computed by subtracting \coltimetrain from \coltimecommonstep) is still clear (16.5\%). For step size 10, \bmcsdslexp and \kissat unroll to the same bound on each instance, but it takes \bmcsdslexp 12.7\% less time to get there. It is not too surprising that the performance gain resulting from \bmcsdsldev is smaller than from \bmcsdslexp. The smaller strategy space has far fewer strategy options and might simply not contain a better strategy than the default one.

In \extref{App.~\ref{sec:other-sys}}, we also consider two additional \sys configurations. One includes all boolean flags in the strategy space; the other uses local-search-based tuning instead of machine learning to pick the solving strategy. Both configurations perform worse than the \kissat-based BMC baseline. It is worth noting that local-search-based tuning does also result in speedup in \coltimecommonstep and improves upon \kissat on 16 of the 24 instances. However, the performance gain is less significant compared to \bmcsdsldev and, on certain benchmarks, tuning landed on parameter settings that drastically harm the performance. This suggests that using empirical performance models can be more robust than direct tuning in our setting.

\subsection{Mini Hardware Model Checking Competition}

We evaluate \bmcsdslexp with step size 10 on all the satisfiable and unsolved bitvector benchmarks from HWMCC. As in the competition, we use a time limit of 1 hour for this experiment. We consider the basic \kissat-based BMC procedure (also using step size 10) as a baseline. In addition, we perform an apples-to-oranges comparison to two algorithm portfolios, one of \pono and the other of the \avr model checker~\cite{goel2020avr}, which was the winner of the most recent competition.\unskip
\footnote{We hope to also compare with the bit-level solver \abc~\cite{brayton2010abc} but have no information about the commands and version used for the competition. We have contacted the \abc team and will include such results after hearing back.}
We use the competition portfolio of \avr, which consists of 16 single-threaded solving modes. The \pono portfolio contains 13 single-threaded modes selected by the developers of \pono. Each mode can construct counter-examples. The \avr portfolio contains two BMC modes, both with step size 5. The \pono portfolio contains 1 BMC mode, with step size 11. 

The number of solved instances and the total time on solved instances are shown in Tab.~\ref{tab:hwmcc}. To study the complementarity of the configurations, we also report the number of unique solutions and the performance of a virtual best configuration. Results on individual benchmarks are reported in \extref{App.~\ref{sec:raw-hwmcc}}.

\bmcsdslexp solves all the instances solved by \kissat plus 7 more, suggesting that while \sys might create overhead for easy instances, this overhead is overcome by benefits in the long run. Impressively, those 7 problems are also not solved by the \avr and \pono portfolios. This suggests that including an \sys-driven BMC procedure in a model checking algorithm portfolio can be beneficial.

\input{results/hwmcc}

\subsection{Ablation studies of training budget and model architecture}

\begin{figure}[t]
    \centering
    \includegraphics[width=0.94\linewidth]{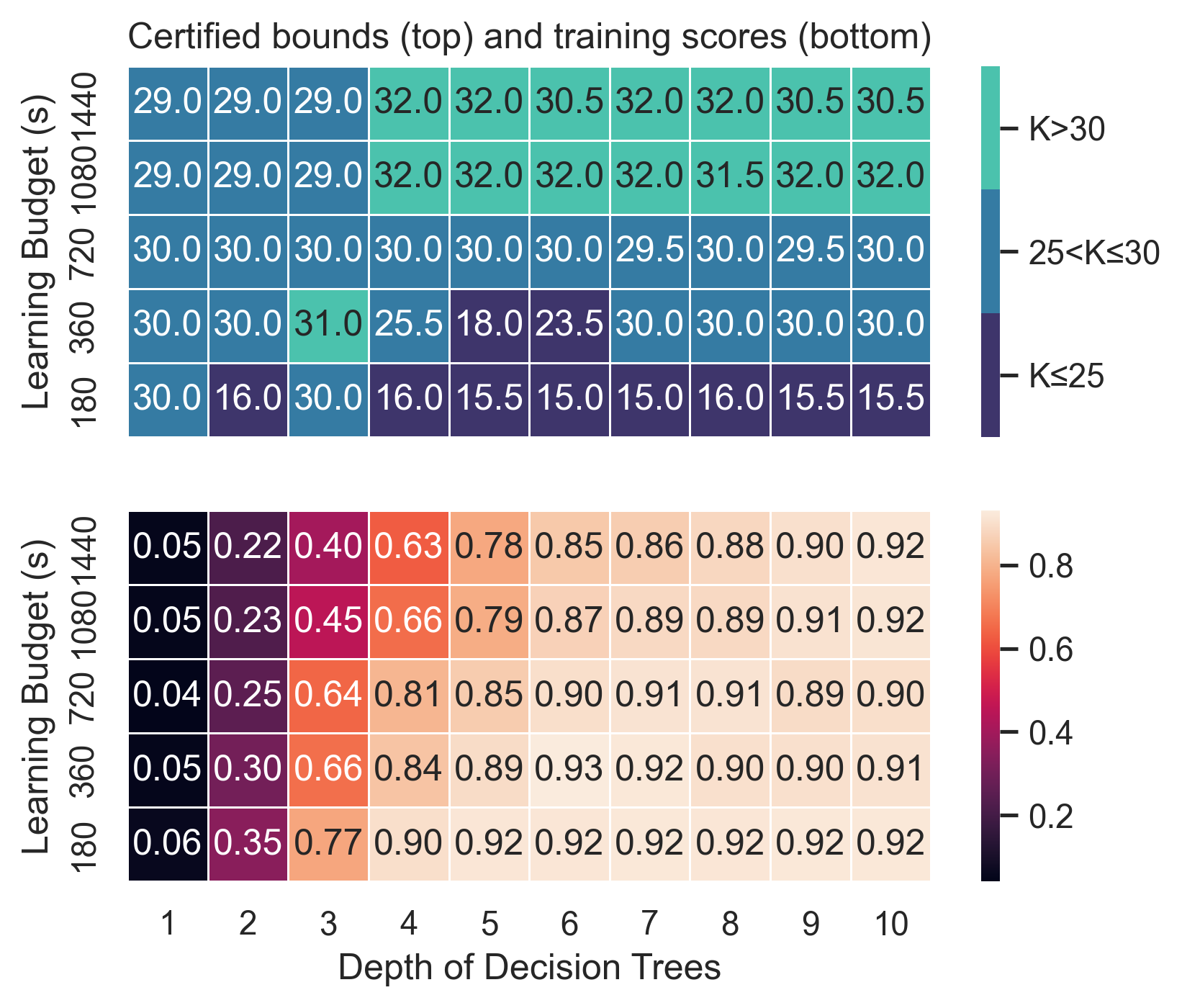}
    \caption{Varying the learning budget and tree depth on \benchmark{picor.pcregs-p0}.}
    \label{fig:ablation}
\end{figure}

To study the effect of dataset size and model accuracy, we select one benchmark \benchmark{picor.pcregs-p0}, and vary the learning budget (in seconds) in the set \set{180,360, 720,1080, 1440} and the decision tree depth from 1 to 10.\footnote{For this experiment we use a fixed tree depth instead of the dynamic one described in Sec.~\ref{subsec:learn}.}
We consider all 50 combinations of the two. For each combination, we run \bmcsdslexp (step size 1, time limit 2 hours) 12 times, each time with a different random seed ($0\ldots11$); we show the median certified bound in the top half of Fig.~\ref{fig:ablation} and the average training score ($R^2$ score, the larger the better) in the last learning epoch in the bottom half. The largest bound certified by \kissat without \sys is 30.

Noticeably, on this instance, improvements in the certified bound are achieved when the depth of the tree is at least 4 and the learning budget is at least 1080 seconds. This suggests that both a sufficient amount of training data and an accurate model are necessary for \sys to work in practice. If not enough data is collected (bottom right), the machine learning model cannot extrapolate well to new problem instances. On the other hand, if the machine learning model is not accurate enough (top left), the strategy it suggests can also be misleading. Determining the optimal learning budget on a per-benchmark basis is a topic worth studying in the future.

%% file: results/unsolved_sdsl_vs_kissat.tex
\begin{table*}[t]
\setlength\tabcolsep{7.5pt}
\centering
\caption{Evaluation of \bmcsdslexp on BV benchmarks of Hardware Model Checking Competition 2020 that were not solved during the competition. 
\coliters is the number of training epochs. \coltimetrain is the time spent on data collection, training, and inference. \coltimecommonstep is the cumulative time (\coltimetrain \textbf{included}) to solve all the formulas up until the largest bound commonly solved by \bmcsdslexp\ and \kissat. %For \bmcsdslexp\ , the remaining time ($7200 - \coltimecommonstep$) is spent on solving larger bounds (if any) beyond the commonly reached bound. 
\colbound is the largest solved bound within 2 hours and is \unsafe{highlighted} if a violation is found (i.e., the benchmark is solved).
} 
\small
\label{tab:eval-exp}
\setlength\tabcolsep{5.5pt}
\begin{tabular}{lcccccccccccccc}
\toprule
& \multicolumn{7}{c}{step size = 1} & \multicolumn{7}{c}{step size = 10} \\
\cmidrule(lr){2-8} \cmidrule(lr){9-15} 
 & \multicolumn{4}{c}{\tabtitle{\bmcsdslexp}} & \multicolumn{2}{c}{\tabtitle{\kissat}} & \multicolumn{1}{c}{\tabtitle{\pono}} & \multicolumn{4}{c}{\tabtitle{\bmcsdslexp}} & \multicolumn{2}{c}{\tabtitle{\kissat}} & \multicolumn{1}{c}{\tabtitle{\pono}} \\
\cmidrule(lr){2-5} \cmidrule(lr){6-7} \cmidrule(lr){8-8} \cmidrule(lr){9-12} \cmidrule(lr){13-14} \cmidrule(lr){15-15} 
\tabtitle{Benchmark}
& \coliters & \coltimetrain & \coltimecommonstep & \colbound & \coltimecommonstep & \colbound & \colbound & \coliters & \coltimetrain & \coltimecommonstep & \colbound & \coltimecommonstep & \colbound & \colbound \\
\benchmark{arb.n2.w128.d64} & 10 & 1040 & \best{4816} & \best{59} & 7198 & 54 & 45 & 2 & 751 & \best{3676} & \unsafe{70} & 4940 & \unsafe{70} & 50 \\
\benchmark{arb.n2.w64.d64} & 9 & 936 & \best{4431} & \best{59} & 6515 & 53 & 48 & 2 & 638 & \best{3985} & \unsafe{70} & 4224 & \unsafe{70} & 50 \\
\benchmark{arb.n2.w8.d128} & 10 & 1031 & \best{5599} & \best{54} & 6795 & 52 & 45 & 2 & 1127 & \best{3932} & 60 & 5528 & 60 & 50 \\
\benchmark{arb.n3.w16.d128} & 9 & 937 & \best{4898} & \best{58} & 7118 & 53 & 48 & 2 & 807 & \best{3653} & \best{70} & 5792 & 60 & 60 \\
\benchmark{arb.n3.w64.d128} & 10 & 1062 & \best{4856} & \best{58} & 7132 & 53 & 44 & 1 & 84 & \best{3427} & 60 & 5120 & 60 & 50 \\
\benchmark{arb.n3.w64.d64} & 10 & 985 & \best{4809} & \best{58} & 7055 & 53 & 46 & 2 & 919 & \best{3736} & \unsafe{70} & 5018 & \unsafe{70} & 50 \\
\benchmark{arb.n3.w8.d128} & 9 & 983 & \best{4760} & \best{58} & 6996 & 53 & 46 & 2 & 906 & \best{3227} & 60 & 6017 & 60 & 50 \\
\benchmark{arb.n4.w128.d64} & 9 & 963 & 6143 & \best{54} & \best{6107} & 53 & 46 & 2 & 1086 & \best{3704} & \unsafe{70} & 5627 & \unsafe{70} & 50 \\
\benchmark{arb.n4.w16.d64} & 10 & 984 & \best{4768} & \best{57} & 6599 & 53 & 49 & 2 & 580 & \best{3349} & \unsafe{70} & 5231 & \unsafe{70} & \unsafe{70} \\
\benchmark{arb.n4.w8.d64} & 10 & 1081 & \best{5863} & \best{54} & 6659 & 52 & 48 & 2 & 617 & \best{3177} & \unsafe{70} & 4563 & \unsafe{70} & 50 \\
\benchmark{arb.n5.w128.d64} & 9 & 1105 & \best{5095} & \best{57} & 6731 & 53 & 47 & 1 & 139 & \best{4204} & \unsafe{70} & 5587 & \unsafe{70} & 50 \\
\benchmark{circ.w128.d128} & 7 & 845 & \best{5746} & \best{46} & 6549 & 44 & 39 & 1 & 536 & 2348 & 30 & \best{1743} & 30 & 20 \\
\benchmark{circ.w128.d64} & 8 & 887 & \best{6071} & \best{47} & 6452 & 46 & 39 & 1 & 747 & 2688 & 30 & \best{1962} & 30 & 20 \\
\benchmark{circ.w16.d128} & 11 & 967 & \best{4997} & \best{58} & 7179 & 54 & 47 & 2 & 876 & 5778 & 50 & \best{4978} & 50 & 40 \\
\benchmark{circ.w64.d128} & 9 & 870 & \best{5409} & \best{51} & 6895 & 48 & 44 & 1 & 232 & 4575 & 40 & \best{4294} & 40 & 30 \\
\benchmark{dspf.p22} & 4 & 1078 & \best{3741} & \best{31} & 5654 & 28 & 27 & 1 & 285 & 289 & 20 & \best{90} & 20 & 20 \\
\benchmark{pgm.3.prop5} & 10 & 1019 & 7128 & 128 & \best{5663} & \best{133} & 131 & 2 & 618 & 6844 & 190 & \best{6052} & 190 & 170 \\
\benchmark{picor.AX.nom.p2} & 2 & 929 & \best{3307} & \best{16} & 3636 & 15 & 14 & 1 & 117 & 279 & \unsafe{20} & \best{151} & \unsafe{20} & \unsafe{20} \\
\benchmark{picor.pcregs-p0} & 5 & 992 & \best{4020} & \best{32} & 6402 & 30 & 30 & 0 & 0 & 83 & 20 & 85 & 20 & 20 \\
\benchmark{picor.pcregs-p2} & 5 & 840 & 6149 & 30 & \best{5003} & \best{31} & 31 & 0 & 0 & 89 & 20 & 91 & 20 & 20 \\
\benchmark{shift.w128.d64} & 7 & 858 & \best{4098} & \best{27} & 5393 & 25 & 21 & 1 & 582 & 678 & \best{30} & \best{353} & 20 & 20 \\
\benchmark{shift.w16.d128} & 9 & 1084 & \best{3020} & \best{45} & 5817 & 39 & 28 & 2 & 864 & \best{1731} & \best{50} & 2606 & 40 & 20 \\
\benchmark{shift.w32.d128} & 8 & 821 & \best{2778} & \best{43} & 6273 & 35 & 27 & 1 & 216 & 2656 & 30 & \best{2424} & 30 & 20 \\
\benchmark{zipversa.p03} & 5 & 1110 & \best{2961} & \best{82} & 6683 & 59 & 45 & 2 & 1182 & \best{2138} & 110 & 6622 & 90 & \best{330} \\
\midrule
Mean & 8.1 & 975 & \best{4811} & \best{52.6} & 6354 & 48.7 & 43.1 & 1.5 & 580 & \best{2927} & \best{57.5} & 3712 & 55.4 & 55.4  \\
\bottomrule
\end{tabular}
\end{table*}

%% file: results/unsolved_sdsl_vs_kissat_biere.tex
\begin{table*}[ht]
\centering
\caption{Evaluation of \bmcsdsldev. The setup is the same as Tab.~\ref{tab:eval-exp}}.
\small
\label{tab:eval-dev}
\begin{tabular}{lcccccccccccc}
\toprule
& \multicolumn{6}{c}{step size = 1} & \multicolumn{6}{c}{step size = 10} \\
\cmidrule(lr){2-7} \cmidrule(lr){8-13} 
 & \multicolumn{4}{c}{\tabtitle{\bmcsdsldev}} & \multicolumn{2}{c}{\tabtitle{\kissat}} & \multicolumn{4}{c}{\tabtitle{\bmcsdsldev}} & \multicolumn{2}{c}{\tabtitle{\kissat}} \\
\cmidrule(lr){2-5} \cmidrule(lr){6-7} \cmidrule(lr){8-11} \cmidrule(lr){12-13} 
\tabtitle{Benchmark} & \coliters & \coltimetrain & \coltimecommonstep & \colbound & \coltimecommonstep & \colbound & \coliters & \coltimetrain & \coltimecommonstep & \colbound & \coltimecommonstep & \colbound \\
\benchmark{arb.n2.w128.d64} & 11 & 861 & \best{6714} & 54 & 7198 & 54 & 2 & 631 & \best{3720} & \unsafe{70} & 4940 & \unsafe{70} \\
\benchmark{arb.n2.w64.d64} & 10 & 896 & \best{6158} & \best{54} & 6515 & 53 & 2 & 575 & 4981 & \unsafe{70} & 4224 & \unsafe{70} \\
\benchmark{arb.n2.w8.d128} & 11 & 1006 & \best{6291} & \best{53} & 6795 & 52 & 2 & 684 & \best{3674} & 60 & 5528 & 60 \\
\benchmark{arb.n3.w16.d128} & 9 & 824 & \best{6061} & \best{54} & 7118 & 53 & 2 & 635 & \best{3905} & 60 & 5792 & 60 \\
\benchmark{arb.n3.w64.d128} & 10 & 966 & \best{6417} & \best{54} & 7132 & 53 & 2 & 735 & \best{3937} & 60 & 5120 & 60 \\
\benchmark{arb.n3.w64.d64} & 11 & 1053 & \best{5452} & \best{55} & 7055 & 53 & 2 & 598 & 5619 & \unsafe{70} & \best{5018} & \unsafe{70} \\
\benchmark{arb.n3.w8.d128} & 9 & 857 & \best{5780} & \best{55} & 6996 & 53 & 2 & 635 & \best{3703} & 60 & 6017 & 60 \\
\benchmark{arb.n4.w128.d64} & 10 & 901 & \best{5859} & \best{55} & 6107 & 53 & 2 & 702 & 6712 & \unsafe{70} & \best{5627} & \unsafe{70} \\
\benchmark{arb.n4.w16.d64} & 10 & 898 & \best{5626} & \best{55} & 6599 & 53 & 2 & 462 & \best{3266} & \unsafe{70} & 5231 & \unsafe{70} \\
\benchmark{arb.n4.w8.d64} & 11 & 973 & \best{4749} & \best{57} & 6659 & 52 & 2 & 532 & \best{3774} & 70 & 4563 & 70 \\
\benchmark{arb.n5.w128.d64} & 10 & 869 & \best{5672} & \best{55} & 6731 & 53 & 1 & 92 & \best{5283} & \unsafe{70} & 5587 & \unsafe{70} \\
\benchmark{circ.w128.d128} & 8 & 694 & 6948 & 43 & \best{5879} & \best{44} & 1 & 400 & 2158 & 30 & \best{1743} & 30 \\
\benchmark{circ.w128.d64} & 8 & 829 & 6650 & 45 & \best{5803} & \best{46} & 1 & 788 & 2663 & 30 & \best{1962} & 30 \\
\benchmark{circ.w16.d128} & 12 & 969 & \best{6961} & 54 & 7179 & 54 & 2 & 433 & \best{4927} & 50 & 4978 & 50 \\
\benchmark{circ.w64.d128} & 10 & 713 & \best{6759} & 48 & 6895 & 48 & 1 & 179 & 4361 & 40 & \best{4294} & 40 \\
\benchmark{dspf.p22} & 4 & 959 & \best{4148} & \best{31} & 5654 & 28 & 1 & 185 & 177 & 20 & \best{90} & 20 \\
\benchmark{pgm.3.prop5} & 16 & 920 & \best{6933} & 133 & 6996 & 133 & 3 & 557 & \best{5664} & 190 & 6052 & 190 \\
\benchmark{picor.AX.nom.p2} & 2 & 590 & \best{3540} & 15 & 3636 & 15 & 1 & 59 & 191 & \unsafe{20} & \best{151} & \unsafe{20} \\
\benchmark{picor.pcregs-p0} & 6 & 873 & 6573 & 28 & \best{3337} & \best{30} & 0 & 0 & 86 & 20 & 85 & 20 \\
\benchmark{picor.pcregs-p2} & 5 & 646 & 6045 & 28 & \best{2786} & \best{31} & 0 & 0 & 86 & 20 & 91 & 20 \\
\benchmark{shift.w128.d64} & 8 & 666 & 5965 & 25 & \best{5393} & 25 & 1 & 1392 & 601 & 20 & \best{353} & 20 \\
\benchmark{shift.w16.d128} & 9 & 940 & \best{5209} & \best{40} & 5817 & 39 & 1 & 139 & \best{2058} & 40 & 2606 & 40 \\
\benchmark{shift.w32.d128} & 8 & 796 & \best{5551} & \best{36} & 6273 & 35 & 1 & 217 & 2615 & 30 & \best{2424} & 30 \\
\benchmark{zipversa.p03} & 2 & 460 & 7037 & 59 & \best{6683} & 59 & 1 & 268 & \best{3650} & 90 & 6622 & 90 \\
\midrule
Mean & 8.8 & 840 & \best{5962} & \best{49.4} & 6134 & 48.7 & 1.5 & 454 & \best{3242} & 55.4 & 3712 & 55.4 \\
\bottomrule
\end{tabular}
\end{table*}

%% file: results/hwmcc.tex
\begin{table}[t]
\centering
\caption{Comparison with two algorithm portfolios on satisfiable and unsolved BV HWMCC benchmarks (89 in total).} 
\small
\label{tab:hwmcc}
\begin{tabular}{lccccc}
\toprule
 \tabtitle{Config.} & \tabtitle{Threads} & \tabtitle{Slv.} & \tabtitle{Time} & \tabtitle{Unique} \\ 
 \cmidrule(lr){2-5}
\bmcsdslexp & 1 & \textbf{68} & 27362 & \textbf{7}\\
\kissat~ & 1 & 61 & 6358 & 0 \\
\avrp~ & 16 & 48 & 12113 & 2\\
\ponop~ & 13 & 63 & 10723 & 0\\
\midrule
\solver{Virtual Best} & 31 & 72 & 24700 & -- \\ 
\bottomrule
\end{tabular}
\end{table}

%% file: 50conclusion.tex
\section{Conclusion, limitations, and future work}
\label{sec:concl}
We introduced Self-Driven Strategy Learning, a conceptually simple, easy-to-implement online learning approach for solving sets of related problems in automated reasoning. We presented the methodology formally as a set of transition rules and instantiated it in the context of Bounded Model Checking. Our experiments show that equipping a BMC-procedure with \sys results in a significant performance boost, both in terms of certified bounds and solved instances, when comparing against state-of-the-art open-source model checkers.

One thing to consider when applying \sys is that a good return on investment in learning depends on a sensible \emph{a priori} choice of the strategy space.
%; a limitation that we plan to address in future work. 
Another limitation is that when the problem set is small, gathering sufficient training data can be challenging. An intriguing question is whether a problem can be decomposed into sub-problems automatically in order to obtain sufficient data. 
Other future directions include alternative orders of applying the \sys rules, applying \sys to other automated reasoning tasks (e.g., symbolic execution, max-satisfiability, iterative abstraction refinement), and combining \sys with offline learning and incremental solving.

\paragraph{Acknowledgments}
We thank the anonymous reviewers for their careful reviews and constructive feedback.  This work was supported in part by the National Science Foundation (grant 1269248) and by the Stanford Center for Automated Reasoning.  Additionally, the NASA University Leadership initiative (grant \#80NSSC20M0163)
provided funds to assist the authors with their research, but this article solely
reflects the opinions and conclusions of its authors and not any NASA entity.

%% file: 90appendix.tex
\newpage
\onecolumn
\appendices

\section{Versions of the solvers used in the BMC implementation and the experimental evaluation}
\label{sec:versions}

\begin{table}[ht]
\centering
\begin{tabular}{ll}
\toprule
\tabtitle{Solver} & \tabtitle{Version} \\
\midrule
\kissat & \url{https://github.com/arminbiere/kissat/blob/97917ddf2} \\
\pono for generating SMTLIB files &
\url{https://github.com/anwu1219/pono/blob/ed11ef3eb} \\
\boolector &
\url{https://github.com/Boolector/boolector/blob/95859db82} \\
\pono for comparison &
\url{https://github.com/upscale-project/pono/blob/8b2a94649} \\
\avr &
\url{https://github.com/aman-goel/avr/blob/4cbceda5b} \\
\cadical &
\url{https://github.com/arminbiere/cadical/tree/a5f15211d} \\
\bottomrule
\end{tabular}
\end{table}

\section{Other \sys configurations}
\label{sec:other-sys}

We also consider two additional \sys configurations. \bmcsdslall is the same as \bmcsdslexp and \bmcsdsldev, except it considers a strategy space of all 30 boolean flags of \kissat. 
\bmcsdslexptune operates on the larger strategy space \strategyspaceexp and performs online tuning on the first $p$ BMC problems. Using a fixed $p$ for all benchmarks does not do this approach justice. Rather, $p$ is decided dynamically: \bmcsdslexptune keeps solving the BMC problems and keep track of the total solving time $T$; when $\paramnumsamples \cdot T$ exceeds the sampling budget $\paramstopthreshold$ for solving the current problem index \step, it tunes on all the problems seen before \step using the conditional sampling procedure described in Sec.~\ref{sec:method2}. We have also tried other local search methods instead of the M-H algorithm, but the effect is similar. 

The performance of these two configurations on the unsolved benchmarks is shown in Tab.~\ref{tab:eval-other}. The step size is 1. \bmcsdslall performs significantly worse than \kissat. This is not a surprise considering we are only allowed to draw 100 samples from $2^{30}$ possible candidates each time, which is not sufficient to capture the dependencies between parameter values or even land on any effective solving strategies. This suggests that the strategy space cannot be too large in the online learning setting. On the other hand, \bmcsdslexptune does result in speedup in \coltimecommonstep and improve upon \kissat on 16 of the 24 instances. However, the performance gain is less significant compared to \bmcsdsldev (Tab.~\ref{tab:eval-exp}). Moreover, the average certified bound by \bmcsdslexptune is smaller than \kissat, because the tuning appears to land on particularly harmful parameter settings on certain benchmarks (e.g., \benchmark{picor.pcregs-p0}). This suggests that using empirical performance model (e.g., a Random Forest) is more robust than direct tuning in the online meta-algorithmic design setting.

\input{results/unsolved_other_configs}

\clearpage
\section{Cross-examination of learned solving strategies}

We study the effect of applying the learned parameter setting on one set of problems to other sets. In particular, we identify five parameter settings learned by \bmcsdslexp that have pairwise hamming distances of at least 6. For each of the five corresponding benchmarks, we identify the largest certified bound, and solved the corresponding SAT formula with \kissat using each of the five parameter settings. A CPU timeout of 1800 seconds is given.

% [3, 9, 13, 16, 19]

The benchmarks being considered and the learned parameter settings (only non-default values shown) are:
\begin{enumerate}
    \item \benchmark{arb.n3.w64.d128}: {ands}=0, {chrono}=0, {eliminateint}=50, {forwardeffort}=200, {ifthenelse}=0, {probeint}=10, {stable}=0, {substituteeffort}=20, {vivifyeffort}=200
    \item \benchmark{circ.w128.d128}: {ands}=0, {bumpreasonsrate}=1, {chrono}=0, {eliminateint}=50, {eliminateocclim}=20, {stable}=0, {vivifyeffort}=200
    \item \benchmark{pgm.3.prop5}: {ands}=0, {forwardeffort}=200, {stable}=0, {subsumeocclim}=10, {vivifyeffort}=200
    \item \benchmark{picor.pcregs-p2}: {eliminateocclim}=20, {forwardeffort}=200, {ifthenelse}=0
    \item \benchmark{shift.w32.d128}: {bumpreasonsrate}=1, {eliminateint}=50, {eliminateocclim}=20, {forwardeffort}=200, {probeint}=10, {stable}=0, {substituteeffort}=20, {subsumeocclim}=10, {vivifyeffort}=200
\end{enumerate}

The result is shown below:

\begin{table}[h]
\centering
\begin{tabular}{lccccc}
\toprule
 & \tabtitle{Problem 1} & \tabtitle{Problem 2}  & \tabtitle{Problem 3} & \tabtitle{Problem 4}  & \tabtitle{Problem 5} \\ 
 \cmidrule(lr){2-6}
\tabtitle{Strategy 1} &     445  &     978  &     210  &    --  &    -- \\
\tabtitle{Strategy 2} &     \textbf{405}  &     \textbf{637}  &  241  &    --  &    1357 \\
\tabtitle{Strategy 3} &     507  &     761  &     \textbf{188}  &    --  &     -- \\
\tabtitle{Strategy 4} &     1107  &     864  &     256  &    \textbf{1434}  &    -- \\
\tabtitle{Strategy 5} &     434  &     817  &     215  &    --  &     \textbf{968}  \\
\bottomrule
\end{tabular}
\end{table}

The strategy learned specifically for a benchmark generally performs better than the alternative strategies. In particular, Problem 4 can be solved within the time limit only using Strategy 4. On the other hand, solving Problem 5 using Strategy 4 results in a timeout. This suggests that the optimal strategies for different instances might contradict with one another.

\clearpage
\section{Effect of Incremental Solving}
\label{sec:incremental}

We hoped to also study the interplay between \sys and incremental solving. However, the effect of incremental solving is mixed on the 24 unsolved bit-vector benchmarks from HWMCC. We focus on the state-of-the-art incremental SAT solver \cadical, which is used by \pono under-the-hood to solve the set of problems from a BMC run. We compare \pono with the BMC procedure we implement using the non-incremental mode of \cadical. The bounds certified by running BMC with step size 1 for 2 hours are shown in Tab.~\ref{tab:cadical}. While \pono, which runs the incremental mode of \cadical, certifies a larger bound on barely more than half of the benchmarks, the non-incremental mode of \cadical can certify larger bounds overall. This experiment suggests that it is worth revisiting BMC-specific incremental solving techniques~\cite{strichman2004accelerating} before investigating the interplay between \sys and incremental solving, which we do believe is an important topic.

\begin{table}[ht]
\centering
\caption{The certified bounds by \pono and a \cadical-based BMC procedure.} 
\small
\label{tab:cadical}
\begin{tabular}{lcc}
\toprule
\tabtitle{Benchmark} & \multicolumn{1}{c}{\tabtitle{\pono}} & \multicolumn{1}{c}{\tabtitle{\cadical}} \\
\midrule
\benchmark{arb.n2.w128.d64} & \best{45} & 44 \\
\benchmark{arb.n2.w64.d64} & \best{48} & 45 \\
\benchmark{arb.n2.w8.d128} & 45 & 45 \\
\benchmark{arb.n3.w16.d128} & \best{48} & 44 \\
\benchmark{arb.n3.w64.d128} & 44 & \best{45} \\
\benchmark{arb.n3.w64.d64} & \best{46} & 44 \\
\benchmark{arb.n3.w8.d128} & \best{46} & 45 \\
\benchmark{arb.n4.w128.d64} & \best{46} & 44 \\
\benchmark{arb.n4.w16.d64} & \best{49} & 44 \\
\benchmark{arb.n4.w8.d64} & \best{48} & 45 \\
\benchmark{arb.n5.w128.d64} & \best{47} & 44 \\
\benchmark{circ.w128.d128} & 39 & \best{44} \\
\benchmark{circ.w128.d64} & 39 & \best{43} \\
\benchmark{circ.w16.d128} & 47 & 47 \\
\benchmark{circ.w64.d128} & 44 & \best{45} \\
\benchmark{dspf.p22} & 27 & \best{34} \\
\benchmark{pgm.3.prop5} & 131 & 131 \\
\benchmark{picor.AX.nom.p2} & 14 & \best{15} \\
\benchmark{picor.pcregs-p0} & \best{30} & 27 \\
\benchmark{picor.pcregs-p2} & \best{31} & 27 \\
\benchmark{shift.w128.d64} & 21 & \best{22} \\
\benchmark{shift.w16.d128} & 28 & \best{39} \\
\benchmark{shift.w32.d128} & 27 & \best{33} \\
\benchmark{zipversa.p03} & \best{45} & 43 \\
\midrule
Mean & 43.1 & 43.3 \\
\bottomrule
\end{tabular}
\end{table}

\newpage
\section{Executions of all BMC configurations with step size 1 on the unsolved bitvector HWMCC benchmarks}
\label{sec:raw}

\begin{figure}[ht]
    \centering
\foreach \i in {0,...,11} {
  \includegraphics[width=0.41\linewidth]{figs/execution_cactus_step1_all_configs/\i.png}
}
\end{figure}
\begin{figure}[ht]
    \centering
\foreach \i in {12,...,23} {
  \includegraphics[width=0.41\linewidth]{figs/execution_cactus_step1_all_configs/\i.png}
}
\end{figure}

\newpage
\section{Duration of each training epoch}
\label{sec:training-ep}

The duration (in seconds) of each training epoch during the run of \bmcsdslexp on the unsolved bitvector HWMCC benchmarks with step size 1 is reported in the table below:

\begin{table}[ht]
\centering
\small
\begin{tabular}{lc}
\toprule
\tabtitle{Benchmark} & \tabtitle{$ep_1$, $ep_2$, \ldots} \\
\midrule
\benchmark{arb.n2.w128.d64} & 6, 12, 21, 34, 56, 87, 120, 171, 205, 296 \\
\benchmark{arb.n2.w64.d64} & 12, 17, 30, 44, 76, 132, 147, 192, 255 \\
\benchmark{arb.n2.w8.d128} & 4, 12, 17, 29, 46, 84, 112, 175, 219, 300 \\
\benchmark{arb.n3.w16.d128} & 10, 18, 35, 63, 86, 133, 162, 253, 143 \\
\benchmark{arb.n3.w64.d128} & 6, 12, 25, 39, 76, 119, 145, 233, 146, 223  \\
\benchmark{arb.n3.w64.d64} & 6, 11, 22, 38, 56, 93, 130, 185, 246, 162 \\
\benchmark{arb.n3.w8.d128} & 10, 17, 36, 51, 81, 129, 216, 241, 171 \\
\benchmark{arb.n4.w128.d64} & 10, 14, 33, 50, 71, 144, 166, 184, 259 \\
\benchmark{arb.n4.w16.d64} & 6, 10, 19, 36, 56, 88, 136, 198, 251, 150 \\
\benchmark{arb.n4.w8.d64} & 6, 10, 19, 35, 53, 82, 107, 180, 239, 315 \\
\benchmark{arb.n5.w128.d64} & 12, 20, 34, 50, 81, 140, 187, 247, 304 \\
\benchmark{circ.w128.d128} & 5, 15, 35, 48, 122, 220, 373 \\
\benchmark{circ.w128.d64} & 5, 15, 25, 65, 129, 169, 286, 161 \\
\benchmark{circ.w16.d128} & 2, 6, 13, 25, 47, 80, 110, 137, 217, 200, 94 \\
\benchmark{circ.w64.d128} & 3, 6, 11, 22, 50, 82, 142, 219, 300 \\
\benchmark{dspf.p22} & 178, 286, 274, 317 \\
\benchmark{pgm.3.prop5} & 55, 73, 85, 106, 109, 85, 83, 111, 200, 78 \\
\benchmark{picor.AX.nom.p2} & 102, 770 \\
\benchmark{picor.pcregs-p0} & 21, 68, 242, 360, 275 \\
\benchmark{picor.pcregs-p2} & 21, 103, 180, 312, 200, 48 \\
\benchmark{shift.w128.d64} &  5, 12, 29, 94, 160, 203, 327 \\
\benchmark{shift.w16.d128} & 2, 7, 24, 55, 80, 118, 176, 349, 239 \\
\benchmark{shift.w32.d128} & 3, 7, 32, 74, 89, 152, 236, 197\\
\benchmark{zipversa.p03} &  40, 146, 208, 302, 398 \\
\bottomrule
\end{tabular}
\end{table}

\clearpage
\section{Raw Results on the satisfiable and unknown bit-vector benchmarks from HWMCC'20}
\label{sec:raw-hwmcc}
\textbf{Note}: for \ponop and \avrp, the status is \textsc{MEMOUT} if and only if none of the threads in the portfolio solve the problem and at least one of the threads runs out of the 8GB memory limit.

\begin{table*}[ht]
\setlength\tabcolsep{12pt}
\centering
\begin{tabular}{lllrr}
\toprule
\tabtitle{Benchmark} &       \tabtitle{Configuration} & \tabtitle{Status} &  \tabtitle{Time} &   \tabtitle{Memory} \\
\midrule
\benchmark{anderson.3.prop1-back-serstep.btor2} & \bmcsdslexp &     \textsc{SAT} &       2.3 &    74.3 \\
\benchmark{arbitrated\_top\_n2\_w128\_d32\_e0.btor2} & \bmcsdslexp &     \textsc{SAT} &     227.3 &   353.3 \\
\benchmark{arbitrated\_top\_n2\_w128\_d64\_e0.btor2} & \bmcsdslexp &     \textsc{SAT} &    3336.7 &  1110.0 \\
\benchmark{arbitrated\_top\_n2\_w64\_d64\_e0.btor2} & \bmcsdslexp &     \textsc{SAT} &    2566.7 &   659.1 \\
\benchmark{arbitrated\_top\_n2\_w8\_d128\_e0.btor2} &  \bmcsdslexp &     \textsc{TIMEOUT} &    3600.0 &   402.9 \\
\benchmark{arbitrated\_top\_n2\_w8\_d16\_e0.btor2} & \bmcsdslexp &     \textsc{SAT} &      27.3 &    73.4 \\
\benchmark{arbitrated\_top\_n2\_w8\_d32\_e0.btor2} & \bmcsdslexp &     \textsc{SAT} &     131.3 &   108.6 \\
\bottomrule
\end{tabular}
\end{table*}
\begin{table*}[ht]
\setlength\tabcolsep{12pt}
\centering
\begin{tabular}{lllrr}
\toprule
\tabtitle{Benchmark} &       \tabtitle{Configuration} & \tabtitle{Status} &  \tabtitle{Time} &   \tabtitle{Memory} \\
\midrule
\benchmark{picorv32\_mutBY\_nomem-p7.btor} & \bmcsdslexp &     \textsc{SAT} &      50.7 &   350.5 \\
\benchmark{picorv32\_mutCX\_nomem-p0.btor} & \bmcsdslexp &     \textsc{SAT} &     114.4 &   458.1 \\
\benchmark{picorv32\_mutCX\_nomem-p8.btor} & \bmcsdslexp &     \textsc{SAT} &     253.5 &   454.7 \\
\benchmark{picorv32\_mutCY\_nomem-p0.btor} & \bmcsdslexp &     \textsc{SAT} &     102.9 &   360.7 \\
\benchmark{picorv32\_mutCY\_nomem-p3.btor} & \bmcsdslexp &     \textsc{SAT} &      98.2 &   364.0 \\
\benchmark{rast-p03.btor} & \bmcsdslexp &     \textsc{SAT} &       1.5 &    68.9 \\
\benchmark{rast-p06.btor} & \bmcsdslexp &     \textsc{SAT} &       1.6 &    69.1 \\
\bottomrule
\end{tabular}
\end{table*}
\begin{table*}[ht]
\setlength\tabcolsep{12pt}
\centering
\begin{tabular}{lllrr}
\toprule
\tabtitle{Benchmark} &       \tabtitle{Configuration} & \tabtitle{Status} &  \tabtitle{Time} &   \tabtitle{Memory} \\
\midrule
\benchmark{mul9.btor2} &  \kissat &     \textsc{TIMEOUT} &    3600.0 &   177.4 \\
\benchmark{peg\_solitaire.3.prop1-back-serstep.btor2} &  \kissat &     \textsc{SAT} &      59.6 &   157.2 \\
\benchmark{pgm\_protocol.3.prop5-func-interl.btor2} &  \kissat &     \textsc{TIMEOUT} &    3600.9 &  2771.0 \\
\benchmark{picorv32-pcregs-p0.btor} &  \kissat &     \textsc{TIMEOUT} &    3600.0 &   280.1 \\
\benchmark{picorv32-pcregs-p2.btor} &  \kissat &     \textsc{TIMEOUT} &    3600.0 &   274.2 \\
\benchmark{picorv32\_mutAX\_nomem-p0.btor} &  \kissat &     \textsc{SAT} &     154.0 &   452.6 \\
\benchmark{picorv32\_mutAX\_nomem-p2.btor} &  \kissat &     \textsc{SAT} &     140.0 &   452.6 \\
\bottomrule
\end{tabular}
\end{table*}
\begin{table*}[ht]
\setlength\tabcolsep{12pt}
\centering
\begin{tabular}{lllrr}
\toprule
\tabtitle{Benchmark} &       \tabtitle{Configuration} & \tabtitle{Status} &  \tabtitle{Time} &   \tabtitle{Memory} \\
\midrule
\benchmark{blocks.4.prop1-back-serstep.btor2} &  \ponop &    \textsc{SAT} &     333.3 &   301.0 \\
\benchmark{brp2.2.prop1-func-interl.btor2} &  \ponop &    \textsc{SAT} &    1795.6 &  2202.0 \\
\benchmark{brp2.3.prop1-back-serstep.btor2} &  \ponop &    \textsc{SAT} &      25.5 &   124.2 \\
\benchmark{buggy\_ridecore.btor} &  \ponop &    \textsc{SAT} &     137.7 &   846.6 \\
\benchmark{circular\_pointer\_top\_w128\_d128\_e0.btor2} &  \ponop &   \textsc{MEMOUT} &     252.4 &  8000.0 \\
\benchmark{circular\_pointer\_top\_w128\_d64\_e0.btor2} &  \ponop &   \textsc{MEMOUT} &    1956.7 &  8000.0 \\
\benchmark{circular\_pointer\_top\_w128\_d8\_e0.btor2} &  \ponop &    \textsc{SAT} &       1.1 &    34.2 \\
\bottomrule
\end{tabular}
\end{table*}
\begin{table*}[ht]
\setlength\tabcolsep{12pt}
\centering
\begin{tabular}{lllrr}
\toprule
\tabtitle{Benchmark} &       \tabtitle{Configuration} & \tabtitle{Status} &  \tabtitle{Time} &   \tabtitle{Memory} \\
\midrule
\benchmark{arbitrated\_top\_n3\_w64\_d64\_e0.btor2} &  \avrp &   \textsc{MEMOUT} &    1511.9 &  8000.0 \\
\benchmark{arbitrated\_top\_n3\_w8\_d128\_e0.btor2} &  \avrp &   \textsc{MEMOUT} &    1056.6 &  8000.0 \\
\benchmark{arbitrated\_top\_n3\_w8\_d16\_e0.btor2} &  \avrp &    \textsc{SAT} &       7.9 &    60.5 \\
\benchmark{arbitrated\_top\_n4\_w128\_d16\_e0.btor2} &  \avrp &    \textsc{SAT} &      11.1 &   131.8 \\
\benchmark{arbitrated\_top\_n4\_w128\_d64\_e0.btor2} &  \avrp &   \textsc{MEMOUT} &    1800.1 &  8000.0 \\
\benchmark{arbitrated\_top\_n4\_w16\_d16\_e0.btor2} &  \avrp &    \textsc{SAT} &       7.5 &    73.8 \\
\benchmark{arbitrated\_top\_n4\_w16\_d32\_e0.btor2} &  \avrp &    \textsc{SAT} &     675.8 &   251.4 \\
\bottomrule
\end{tabular}
\end{table*}
\begin{table*}[ht]
\setlength\tabcolsep{12pt}
\centering
\begin{tabular}{lllrr}
\toprule
\tabtitle{Benchmark} &       \tabtitle{Configuration} & \tabtitle{Status} &  \tabtitle{Time} &   \tabtitle{Memory} \\
\midrule
\benchmark{shift\_register\_top\_w64\_d8\_e0.btor2} &  \avrp &    \textsc{SAT} &       5.1 &    88.1 \\
\benchmark{shift\_register\_top\_w8\_d128\_e0.btor2} &  \avrp &   \textsc{MEMOUT} &    2138.0 &  8000.0 \\
\benchmark{stack-p1.btor} &  \avrp &    \textsc{SAT} &       0.2 &    29.8 \\
\benchmark{vis\_arrays\_am2901.btor2} &  \avrp &    \textsc{SAT} &       1.4 &   111.9 \\
\benchmark{vis\_arrays\_buf\_bug.btor2} &  \avrp &    \textsc{SAT} &       2.0 &    36.5 \\
\benchmark{zipversa\_composecrc\_prf-p03.btor} &  \avrp &    \textsc{SAT} &      18.7 &   272.9 \\
\bottomrule
\end{tabular}
\end{table*}

\clearpage
\section{Side-by-side results on the satisfiable and unknown bitvector benchmarks from HWMCC'20}

\begin{table*}[ht]
\centering

\begin{tabular}{llrlrlrlr}
\toprule
\tabtitle{Benchmark}  & \multicolumn{2}{c}{\tabtitle{\bmcsdslexp}}  & \multicolumn{2}{c}{\tabtitle{\kissat}}  & \multicolumn{2}{c}{\tabtitle{\ponop}}  & \multicolumn{2}{c}{\tabtitle{\avrp}}\\
\cmidrule(lr){2-3} \cmidrule(lr){4-5} \cmidrule(lr){6-7} \cmidrule(lr){8-9} 
&\tabtitle{Status} &  \tabtitle{Time} & \tabtitle{Status} &  \tabtitle{Time} & \tabtitle{Status} &  \tabtitle{Time} & \tabtitle{Status} &  \tabtitle{Time}\\
\midrule
\benchmark{anderson.3.prop1-back-serstep.btor2} &       \textsc{SAT} &         2.3 &       \textsc{SAT} &         2.4 &       \textsc{SAT} &         0.1 &       \textsc{SAT} &         0.2 \\
\benchmark{arbitrated\_top\_n2\_w128\_d32\_e0.btor2} &       \textsc{SAT} &       227.3 &       \textsc{SAT} &       175.7 &       \textsc{SAT} &       605.6 &       \textsc{SAT} &       455.4 \\
\benchmark{arbitrated\_top\_n2\_w128\_d64\_e0.btor2} &       \textsc{SAT} &      3336.7 &       \textsc{TO} &      3600.0 &       \textsc{MO} &      1785.3 &       \textsc{MO} &      1781.9 \\
\benchmark{arbitrated\_top\_n2\_w64\_d64\_e0.btor2} &       \textsc{SAT} &      2566.7 &       \textsc{TO} &      3600.0 &       \textsc{MO} &      1255.8 &       \textsc{MO} &      1437.3 \\
\benchmark{arbitrated\_top\_n2\_w8\_d128\_e0.btor2} &       \textsc{TO} &      3600.0 &       \textsc{TO} &      3600.0 &       \textsc{MO} &      2957.6 &       \textsc{MO} &      1304.1 \\
\benchmark{arbitrated\_top\_n2\_w8\_d16\_e0.btor2} &       \textsc{SAT} &        27.3 &       \textsc{SAT} &         5.5 &       \textsc{SAT} &         0.7 &       \textsc{SAT} &         6.1 \\
\bottomrule
\end{tabular}
\end{table*}
\begin{table*}[ht!]
\centering
\begin{tabular}{llrlrlrlr}
\toprule
\tabtitle{Benchmark}  & \multicolumn{2}{c}{\tabtitle{\bmcsdslexp}}  & \multicolumn{2}{c}{\tabtitle{\kissat}}  & \multicolumn{2}{c}{\tabtitle{\ponop}}  & \multicolumn{2}{c}{\tabtitle{\avrp}}\\
\cmidrule(lr){2-3} \cmidrule(lr){4-5} \cmidrule(lr){6-7} \cmidrule(lr){8-9} 
&\tabtitle{Status} &  \tabtitle{Time} & \tabtitle{Status} &  \tabtitle{Time} & \tabtitle{Status} &  \tabtitle{Time} & \tabtitle{Status} &  \tabtitle{Time}\\
\midrule
\benchmark{picorv32\_mutBY\_nomem-p6.btor} &       \textsc{SAT} &       118.2 &       \textsc{SAT} &        21.6 &       \textsc{SAT} &        55.3 &       \textsc{MO} &      2104.5 \\
\benchmark{picorv32\_mutBY\_nomem-p7.btor} &       \textsc{SAT} &        50.7 &       \textsc{SAT} &        57.2 &       \textsc{SAT} &       225.8 &       \textsc{MO} &      1926.8 \\
\benchmark{picorv32\_mutCX\_nomem-p0.btor} &       \textsc{SAT} &       114.4 &       \textsc{SAT} &        57.3 &       \textsc{SAT} &       136.9 &       \textsc{TO} &      3600.0 \\
\benchmark{picorv32\_mutCX\_nomem-p8.btor} &       \textsc{SAT} &       253.5 &       \textsc{SAT} &       216.2 &       \textsc{SAT} &       276.9 &       \textsc{MO} &      3140.3 \\
\benchmark{picorv32\_mutCY\_nomem-p0.btor} &       \textsc{SAT} &       102.9 &       \textsc{SAT} &        99.7 &       \textsc{SAT} &        76.8 &       \textsc{MO} &      1899.4 \\
\benchmark{picorv32\_mutCY\_nomem-p3.btor} &       \textsc{SAT} &        98.2 &       \textsc{SAT} &        61.8 &       \textsc{SAT} &        27.4 &       \textsc{MO} &      1931.3 \\
\bottomrule
\end{tabular}
\end{table*}

\clearpage
\section{Scatter plot on the satisfiable and unsolved benchmarks from HWMCC'20}

Below is a scatter plot of the runtime of \bmcsdslexp and \kissat on the satisfiable and unsolved benchmarks from HWMCC'20. Most problems can be solved quickly by both configurations, but \bmcsdslexp can solve a few hard ones in addition. It is also noteworthy that on easy instances, \bmcsdslexp usually does not incur large overhead, either because a counter-example is found immediately, or the reduction in the actual solving time cancels out the extra time spent on learning.

\begin{figure}[ht]
    \centering
    \includegraphics{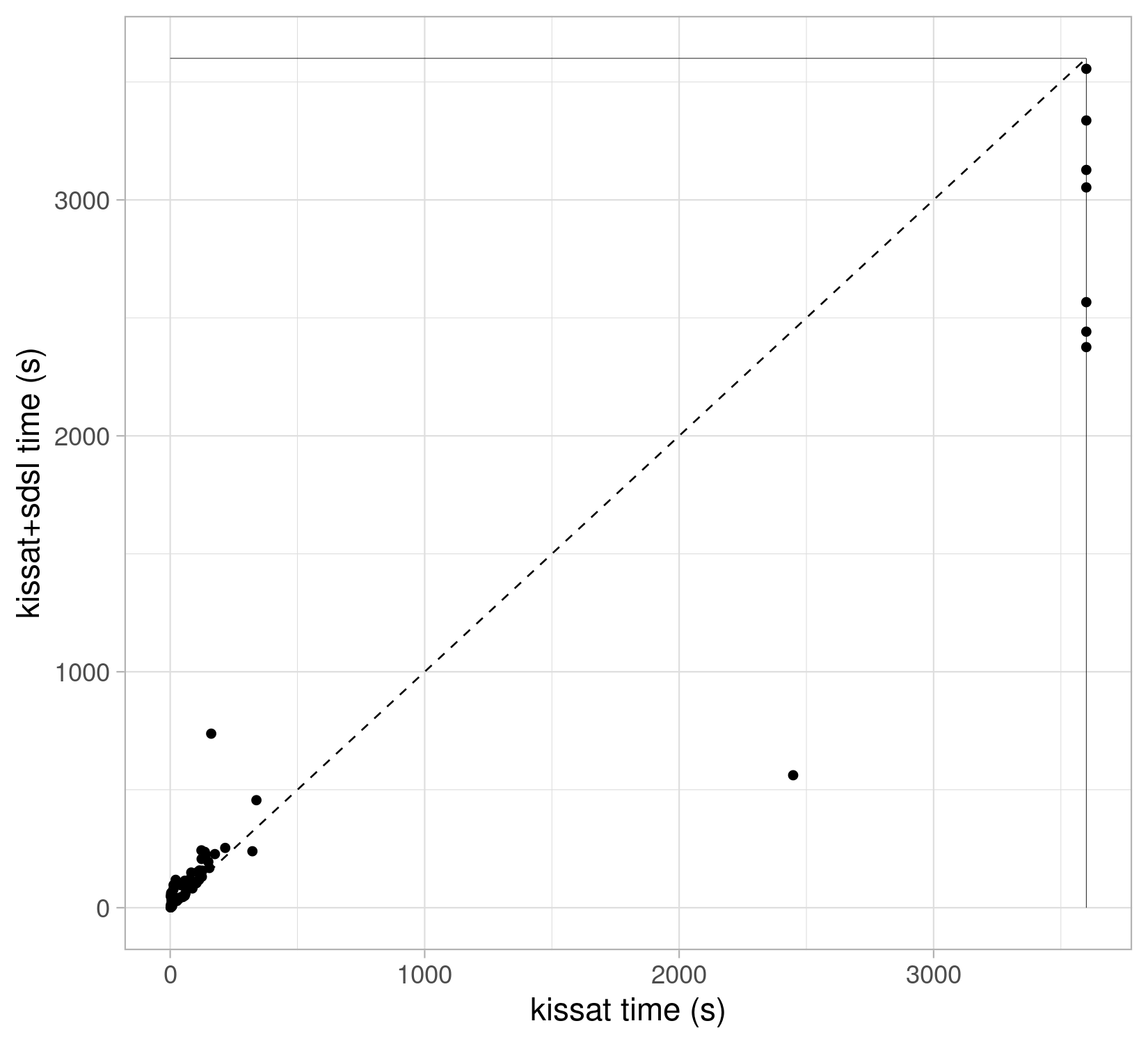}
\end{figure}

\iffalse
\clearpage

\begin{table*}[t]
\setlength\tabcolsep{7.5pt}
\centering

\small
\setlength\tabcolsep{5.5pt}
\begin{tabular}{cccccc}
\toprule
  & \multicolumn{2}{c}{\tabtitle{\kissat+ \sys}} & \multicolumn{2}{c}{\tabtitle{\kissat}} & \tabtitle{\pono} \\
\cmidrule(lr){2-3} \cmidrule(lr){4-5} \cmidrule(lr){6-6}
\tabtitle{Step size} & \tabtitle{Time} & \tabtitle{Bound} & \tabtitle{Time} & \tabtitle{Bound} & \tabtitle{Bound} \\
1 & \best{4811} & \best{52.6} & 6354 & 48.7 & 43.1 \\
10 &  \best{2927} & \best{57.5} & 3712 & 55.4 & 55.4  \\
\bottomrule
\end{tabular}
\end{table*}

\begin{table}[t]
%\setlength\tabcolsep{4.5pt}
\centering
\caption{Comparison with two algorithm portfolios on satisfiable and unsolved BV HWMCC benchmarks (89 in total).} 
\small
\label{tab:hwmcc}
\begin{tabular}{lccccc}
\toprule
 \tabtitle{Config.} & \tabtitle{Threads} & \tabtitle{Slv.} & \tabtitle{Time} & \tabtitle{Unique} \\ 
 \cmidrule(lr){2-5}
\kissat+ \sys & 1 & \textbf{68} & 27362 & \textbf{7}\\
\kissat~ & 1 & 61 & 6358 & 0 \\
\avrp~ & 16 & 48 & 12113 & 2\\
\ponop~ & 13 & 63 & 10723 & 0\\
\midrule
\solver{Virtual Best} & 31 & 72 & 24700 & -- \\ 
\bottomrule
\end{tabular}
\end{table}

\fi

%% file: results/unsolved_other_configs.tex
\begin{table*}[ht]
\centering
\caption{Evaluation of two other \sys configurations \bmcsdslall and \bmcsdslexptune. The setup is the same as Tab.~\ref{tab:eval-exp}}.
\small
\label{tab:eval-other}
\begin{tabular}{lcccccccccccc}
\toprule
& \multicolumn{6}{c}{step size = 1} & \multicolumn{6}{c}{step size = 1} \\
\cmidrule(lr){2-7} \cmidrule(lr){8-13} 
 & \multicolumn{4}{c}{\tabtitle{\bmcsdslall}} & \multicolumn{2}{c}{\tabtitle{\kissat}} & \multicolumn{4}{c}{\tabtitle{\bmcsdslexptune}} & \multicolumn{2}{c}{\tabtitle{\kissat}} \\
\cmidrule(lr){2-5} \cmidrule(lr){6-7} \cmidrule(lr){8-11} \cmidrule(lr){12-13} 
\tabtitle{Benchmark} & \coliters & \coltimetrain & \coltimecommonstep & \colbound & \coltimecommonstep & \colbound & \coliters & \coltimetrain & \coltimecommonstep & \colbound & \coltimecommonstep & \colbound \\
\benchmark{arb.n2.w128.d64} & 9 & 1146 & 7044 & 34 & \best{684} & \best{54} & 1 & 924 & \best{6840} & 54 & 7198 & 54 \\
\benchmark{arb.n2.w64.d64} & 7 & 769 & 6833 & 46 & \best{2908} & \best{53} & 1 & 740 & \best{6342} & \best{54} & 6515 & 53 \\
\benchmark{arb.n2.w8.d128} & 8 & 1326 & 6485 & 51 & \best{5931} & \best{52} & 1 & 1074 & \best{5838} & \best{54} & 6795 & 52 \\
\benchmark{arb.n3.w16.d128} & 8 & 1085 & 6959 & 51 & \best{5764} & \best{53} & 1 & 710 & \best{6779} & 53 & 7118 & 53 \\
\benchmark{arb.n3.w64.d128} & 8 & 1057 & 6929 & 43 & \best{2260} & \best{53} & 1 & 727 & \best{6018} & \best{54} & 7132 & 53 \\
\benchmark{arb.n3.w64.d64} & 9 & 978 & 6963 & 36 & \best{887} & \best{53} & 1 & 799 & \best{6033} & \best{54} & 7055 & 53 \\
\benchmark{arb.n3.w8.d128} & 7 & 928 & 7154 & 35 & \best{879} & \best{53} & 1 & 710 & \best{6673} & 53 & 6996 & 53 \\
\benchmark{arb.n4.w128.d64} & 8 & 1051 & 6630 & 50 & \best{4520} & \best{53} & 1 & 853 & 6563 & 53 & \best{6107} & 53 \\
\benchmark{arb.n4.w16.d64} & 8 & 891 & 6584 & 34 & \best{747} & \best{53} & 1 & 799 & \best{6565} & \best{54} & 6599 & 53 \\
\benchmark{arb.n4.w8.d64} & 7 & 1032 & 6715 & 51 & \best{5953} & \best{52} & 1 & 1282 & \best{6116} & \best{54} & 6659 & 52 \\
\benchmark{arb.n5.w128.d64} & 8 & 974 & 7111 & 50 & \best{4729} & \best{53} & 1 & 695 & \best{6694} & 53 & 6731 & 53 \\
\benchmark{circ.w128.d128} & 7 & 1028 & 6156 & 23 & \best{343} & \best{44} & 1 & 858 & 7192 & 44 & \best{6549} & 44 \\
\benchmark{circ.w128.d64} & 7 & 935 & 6498 & 31 & \best{929} & \best{46} & 1 & 784 & \best{6448} & 46 & 6452 & 46 \\
\benchmark{circ.w16.d128} & 7 & 1068 & 6920 & 39 & \best{1435} & \best{54} & 1 & 1091 & \best{6794} & 54 & 7179 & 54 \\
\benchmark{circ.w64.d128} & 9 & 1034 & 7090 & 41 & \best{3161} & \best{48} & 1 & 638 & \best{4938} & \best{51} & 6895 & 48 \\
\benchmark{dspf.p22} & 3 & 890 & 5614 & 24 & \best{1186} & \best{28} & 1 & 607 & \best{2496} & \best{35} & 5654 & 28 \\
\benchmark{pgm.3.prop5} & 10 & 910 & 7151 & 125 & \best{4980} & \best{133} & 1 & 878 & 6976 & 127 & \best{5330} & \best{133} \\
\benchmark{picor.AX.nom.p2} & 2 & 910 & 1164 & 13 & \best{158} & \best{15} & 1 & 150 & 3876 & 15 & \best{3636} & 15 \\
\benchmark{picor.pcregs-p0} & 5 & 1079 & 6986 & 27 & \best{2427} & \best{30} & 1 & 730 & 20 & 16 & \best{20} & \best{30} \\
\benchmark{picor.pcregs-p2} & 4 & 673 & 5347 & 27 & \best{1942} & \best{31} & 1 & 642 & 5481 & 24 & \best{596} & \best{31} \\
\benchmark{shift.w128.d64} & 7 & 1070 & 7187 & 21 & \best{884} & \best{25} & 1 & 930 & \best{3380} & \best{28} & 5393 & 25 \\
\benchmark{shift.w16.d128} & 7 & 1129 & 6672 & 26 & \best{347} & \best{39} & 1 & 912 & \best{4627} & \best{41} & 5817 & 39 \\
\benchmark{shift.w32.d128} & 7 & 1011 & 7194 & 35 & \best{6273} & 35 & 1 & 1128 & \best{5988} & 35 & 6273 & 35 \\
\benchmark{zipversa.p03} & 3 & 961 & \best{5814} & \best{61} & 6683 & 59 & 1 & 252 & 6424 & 57 & \best{6039} & \best{59} \\
\midrule
Mean & 6.9 & 997 & 6466 & 40.6 & \best{2750} & \best{48.7} & 1 & 788 & \best{5629} & 48.5 & 5864 & \best{48.7} \\
\bottomrule
\end{tabular}
\end{table*}

\iffalse
\begin{table}[ht]
%\setlength\tabcolsep{4.5pt}
\centering
\caption{Summary of results of two additional \sys configurations. \bmcsdslexptune uses tuning instead of learning, and \bmcsdslall is the same as \bmcsdslexp except that it searchs over \emph{all} boolean flags of \kissat.} 
\small
\label{tab:other}
\begin{tabular}{lccc}
\toprule
 \tabtitle{Solver} & \tabtitle{\colbound} & \tabtitle{Speed up in \coltimecommonstep} & \tabtitle{\colbetter} \\ 
 \cmidrule(lr){1-4}
\kissat~ & 48.1 & - & - \\
 \cmidrule(lr){1-4}
\bmcsdslexp & \textbf{52.2} & \textbf{1.33x} & \textbf{19} \\
\bmcsdslexptune & 47.7 & 1.03x & 13 \\
\bmcsdslall & 39.7 & 0.61x & 1 \\
\bottomrule
\end{tabular}
\end{table}
\fi